\pdfoutput=1
\documentclass[11pt]{article}

\usepackage[preprint]{acl}

\usepackage{times}
\usepackage{latexsym}

\usepackage{booktabs}
\usepackage{amsmath} 
\usepackage{rotating} %
\usepackage{multirow} 
\usepackage{arydshln}
\usepackage{xcolor}
\definecolor{deepgreen}{rgb}{0.0, 0.6, 0.0} %
\usepackage{makecell}
\usepackage{subcaption}

\usepackage[T1]{fontenc}

\usepackage[utf8]{inputenc}

\usepackage{microtype}

\usepackage{inconsolata}

\usepackage{graphicx}

\title{Do Large Language Models Excel in Complex Logical Reasoning \\ with Formal Language?}

\author{
    \textbf{Jin Jiang\textsuperscript{1,2\dag}}, 
    \textbf{Jianing Wang\textsuperscript{2\dag}}, 
    \textbf{Yuchen Yan\textsuperscript{2,3}}, 
    \textbf{Yang Liu\textsuperscript{2}}, 
\\
    \textbf{Jianhua Zhu\textsuperscript{1}}, 
    \textbf{Mengdi Zhang\textsuperscript{2}}, 
    \textbf{Xunliang Cai\textsuperscript{2}}, 
    \textbf{Liangcai Gao\textsuperscript{1}}, 
\\
\\
    \textsuperscript{1}Peking University, 
    \textsuperscript{2}Meituan Group, 
    \textsuperscript{3}Zhejiang University, 
\\
\small{
    \textbf{Correspondence:} \href{mailto:jiangjin@stu.pku.edu.cn}{jiangjin@stu.pku.edu.cn}, 
    \href{mailto:gaoliangcai@pku.edu.cn}{gaoliangcai@pku.edu.cn}
}
}

\begin{document}
\maketitle
\begin{abstract}

Large Language Models (LLMs) have been shown to achieve breakthrough performance on complex logical reasoning tasks. 
Nevertheless, most existing research focuses on employing formal language to guide LLMs to derive reliable reasoning paths, while systematic evaluations of these capabilities are still limited.
In this paper, we aim to conduct a comprehensive evaluation of LLMs across various logical reasoning problems utilizing formal languages.
From the perspective of three dimensions, i.e., spectrum of LLMs, taxonomy of tasks, and format of trajectories, our key findings are:
1) \textit{Thinking} models significantly outperform \textit{Instruct} models, especially when formal language is employed;
2) All LLMs exhibit limitations in inductive reasoning capability, irrespective of whether they use a formal language;
3) Data with PoT format achieves the best generalization performance across other languages. 
Additionally, we also curate the formal-relative training data to further enhance the small language models, and the experimental results indicate that a simple rejected fine-tuning method can better enable LLMs to generalize across formal languages and achieve the best overall performance.
Our codes and reports are available at 
\url{https://github.com/jiangjin1999/FormalEval}.

\end{abstract}

\section{Introduction}
Logical reasoning, i.e., deductive, inductive, and abductive, is one of the imperative natural language processing (NLP) tasks and plays a significant role in artificial intelligence (AI) to perform human-like decision-making, task-solving, and deep-thinking~\cite{zhang2021neural, yang2023logical, yu2024natural, xu2025large}.
Different from conventional natural language understanding and generation, logical reasoning requires the AI systems to explicitly provide meticulous elucidation of thoughts and verifiable derivation chains, which is crucial and challenging~\cite {cummins1991conditional}.
Early works have developed multiple formal languages with symbol solvers to make the reasoning steps computable and structured~\cite{ranise2003smt, bulatov2005classifying, bjorner2015nuz}. 

\begin{figure}
    \centering
    \includegraphics[width=1.0\linewidth]{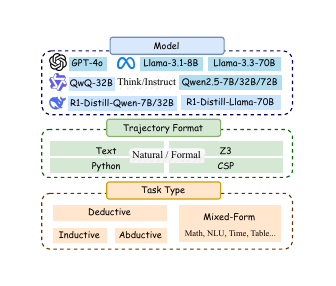}
    \caption{Evaluation framework with three specific dimensions: spectrum of LLMs, taxonomy of logical reasoning tasks, and format of trajectories.}
    \label{fig:preliminary-taxonomy}
\end{figure}

Recently, the emergence of reasoning capabilities in large language models (LLMs) has incentivized significant progress in complex reasoning tasks, such as mathematics, commonsense, and symbol \cite{achiam2023gpt, bi2024deepseek}.
Current studies~\footnote{Due to the space limitations, the detailed description about related works are moved in Appendix~\ref{app:related_work}.} have found that LLMs can achieve remarkable performance with the aid of formal language and symbol solvers, especially when integrating well-designed task-specific instructs~\cite{lyu2023faithful,pan2023logic}, chain-of-thought (CoT) reasoning patterns~\cite{wei2022chain,ye2023satlm}, and valuable solvers' feedback~\cite{he23solving, gao2023pal, wang2024logiclm++}.
Such approaches aim to formalize the given logical problem and constantly adjust the results lean on the solver's feedback.
Despite substantial efforts exhibiting exceptional performance, there are still relatively limited systematic and comprehensive evaluations.
Thus, a natural question remains open: ~\emph{whether the LLM really excels in complex logical reasoning problems with formal language}?

To bridge the gap, this paper endeavors to perform a comprehensive evaluation of LLMs utilizing various formal languages to tackle diverse logical reasoning problems. 
At first, \textbf{we develop the evaluation architecture to clearly express the entire assessment view} (As illustrated in Section~\ref{sec:preliminary}), with the framework shown in Figure \ref{fig:preliminary-taxonomy}.
Specifically, we divide the entire assessment into three distinct dimensions, including the spectrum of LLMs, the taxonomy of logical reasoning tasks, and the format of trajectories.
For the family of LLMs, we further consider different reasoning patterns which has been injected into the model training, such as short thinking (e.g., GPT-4o~\cite{achiam2023gpt}, 
Qwen1.5/2/2.5 ~\cite{bai2023qwen}, 
LLaMA3/3.1/3.3~\cite{grattafiori2024llama}) and long thinking (e.g., DeepSeek-R1-Dsitill-Qwen~\cite{guo2025deepseek}).
For the logical reasoning, we adhere to the classic definitions \cite{flach2000abduction}, categorizing tasks into deductive, inductive, and abductive reasoning. Additionally, we account for tasks that may integrate multiple reasoning types by introducing a new category referred to as mixed-form reasoning.
Regarding the format of trajectories, we consider three main formal languages (``Python'', ``Z3'', ``CSP'') with a default natural language format as ``Text''.

Secondly, we \textbf{perform a thorough evaluation across these three dimensions} (as detailed in Section \ref{sec:part-one}). Many contemporary benchmarks purely emphasize informal text patterns and lack comprehensive integration of different formal languages and logical reasoning tasks \cite{lei2024s3eval,xu2025large, xia2025can}. 
For instance, it is widely recognized that Python is superior to plain text when addressing mathematical problems \cite{friedman2023large, gao2023pal}, but it remains unclear whether Python is equally effective in resolving BBH~\cite{suzgun2022challenging}  and bbeh~\cite{kazemi2025big} problems. To fill this blank, this part aims to investigate whether current LLMs can solve a variety of logical reasoning tasks utilizing different formal languages.
From this study, we derive several intriguing observations: 1) \textit{Thinking} models significantly outperform \textit{Instruct} models, especially when formal language is employed;
2) All LLMs exhibit limitations in inductive reasoning capability, irrespective of whether they use a formal language;
3) LLMs typically produce inferior performance on difficult tasks.
These findings prompt a new inquiry~\emph{Do large models possess generalization capabilities when employing formal languages?}

Thirdly, \textbf{we further investigate the generalization across different reasoning tasks and formal languages} (As illustrated in Section~\ref{sec:part-two}).
To reach this goal, we collect a few training data from the training set of current evaluation tasks, which is classified into three types: deductive, inductive, and abductive.
For each task type, we also provide different trajectories according to the usages of (in)formal languages.
To make a fair comparison, we only use data from a single language type for SFT training, and the training data has the same scale size.
From the experiments, we observe that the LLM can obtain significant in-domain performance on multiple logical reasoning tasks.
In addition, we also discovered an elusive phenomenon that CSP is hard to generalize to other formal and informal languages, but it is easy to generalize from other languages to CSP.
Therefore, we speculate that the poor performance of LLM on some formal languages can be blamed on the lack of pertinent knowledge and potential for stimulated reasoning.

Lastly, based on the previous exploration, 
\textbf{we aim to amplify the capabilities of weaker models in using formal languages to solve reasoning problems.}
Concretely, we propose a simple but effective rejected fine-tuning (RFT) approach to curate different formal-relative training data.
After the enrichment, the overall accuracy of using informal and formal languages for complex logical tasks can be improved by more than 10\%.

In summary, the main contributions are as follows:
\begin{itemize}
    \item In light of the insufficient evaluations of existing works, we aim to collect 66 tasks with multiple widely used formal languages, and provide a comprehensive evaluation for current LLMs across three dimensions, including the spectrum of LLMs, the taxonomy of tasks, and the format of trajectories.

    \item Considering that different formal languages have different expressions for reasoning, we explore the generalization across various formal languages. 

    \item To further enhance the capability of LLMs in utilizing formal languages to solve complex logic reasoning, we introduce a simple but effective rejected fine-tuning method with curated formal-relative data. The experimental results indicate the effectiveness of considering the generalization of formal language across various logical tasks.
\end{itemize}

\section{Preliminary}
\label{sec:preliminary}
As illustrated in Figure~\ref{fig:preliminary-taxonomy}, our evaluation framework is structured along three dimensions: Model, Trajectory Format, and Task Type. In this section, we introduce the two key points of complex reasoning task categorization (Section~\ref{sec:preliminary-one}) and trajectory format design (Section~\ref{sec:preliminary-two}).

\subsection{Taxonomy of Complex Logical Reasoning}
\label{sec:preliminary-one}

Inspired by \citet{xu2025large}, we present a unified taxonomy that categorizes a wide range of complex reasoning tasks into four major types: Deductive, Inductive, Abductive, and Mixed-Form. 
To elaborate, the categorization is based on the nature of reasoning required in human-like thinking in the real world:
1) \textbf{Deductive reasoning} is the forward reasoning process with rules that starts from the given premises to the conclusion~\cite{goel2007anatomy,johnson1999deductive}. Formally, we can denote the process as~\emph{premise}$\stackrel{\text{rule}}{\longrightarrow}$\emph{conclusion}.
2) \textbf{Inductive reasoning} is the process that infers specific rules based on multiple premises and conclusions. It can be represented as (\emph{premise}, \emph{conclusion})$\rightarrow$\emph{rule}.
3) \textbf{Abductive reasoning} is the backward process of deductive which aims to obtain the premise based on conclusion, and the process can be viewed as~\emph{conclusion}$\stackrel{\text{rule}}{\longrightarrow}$\emph{premise}.
4) \textbf{Mix-Form Reasoning} involves at least two of the above three types of reasoning. In real-life scenarios, most complex problems involve mixed reasoning, including but not limited to temporal-spatial reasoning, NLU, knowledge reasoning, and mathematical reasoning.

In pursuit of specific benchmarks based on these categories, we meticulously collect \textbf{66} subsets of data, and the detailed information can be found in Table~\ref{tab:all_dataset}.
The details of the specific datasets are shown in Appendix~\ref{app:datasets}.

\subsection{Trajectory Format}
\label{sec:preliminary-two}

As shown in Figure~\ref{fig:preliminary-taxonomy}, we categorize trajectory formats into two main types: \textbf{informal language} (natural language) and \textbf{formal language}. Informal language can be expressed as free-form text, while formal languages include programming languages (e.g., Python) and logic-based languages (e.g., Z3 and CSP).
They can be modeled as:
\[
\mathcal{LLM}(Q) = \left\langle s_1, s_2, \dots, s_n \right\rangle \xrightarrow{\text{Exec}} A
\]
where \( Q \) is the input question, and \( \mathcal{LLM}(Q) \) represents the trajectory generated by LLM. Each step \( s_i \in \mathcal{L}_{\text{LLM}} \) corresponds to a structured unit (e.g., code or logic expression), and the trajectory is executed by an external engine to produce the final answer \( A \).

For \textbf{PoT}, we use Python 3.12 and its standard library as the execution environment. Each step \( s_i \in \mathcal{LLM}_\text{PoT} \) is a valid Python statement.
For \textbf{Z3}, we adopt the Z3 theorem prover as the executor and Z3 trajectories are composed of declarative symbolic steps \( s_i \in \mathcal{LLM}_\text{Z3} \).
For \textbf{CSP}, we use the \texttt{python-constraint} library as the trajectory executor. Each CSP trajectory \( s_i \in \mathcal{LLM}_\text{CSP} \) consists of variable declarations, domain assignments, and constraint definitions.

In addition, we chose Z3 over Prover9 because Z3 not only supports first-order logic (Prover9-FOL) but also natively supports rich theories such as integers and arrays. More detailed description can be found in section \ref{app:formal-trajectories}.

\section{PART I: Evaluation across LLMs, Tasks, Trajectories}
\label{sec:part-one}

In PART I, we present a comprehensive evaluation across three dimensions: \textbf{Models}, \textbf{Trajectory Formats}, and \textbf{Reasoning Task Types}.
Specifically, we evaluate both \textit{Instruct} and \textit{Thinking} models, ranging from 7B to 72B (see Figure~\ref{fig:preliminary-taxonomy}). For reasoning tasks, we follow the taxonomy introduced in Section~\ref{sec:preliminary-one}. For trajectory formats, we evaluate three formal languages and natural language, as detailed in Section~\ref{sec:preliminary-two}.
All evaluations are conducted in a zero-shot setting. For formal languages (PoT, Z3, CSP), we apply a three-step self-refinement process during code execution. Detailed evaluation settings are provided in Appendix~\ref{app:eval_setups}.

\subsection{Model Performance for Reasoning Tasks and Trajectory Formats}
\label{sec:Model_Performance}

\begin{figure*}[htp]
    \centering
    \includegraphics[width=1\textwidth]{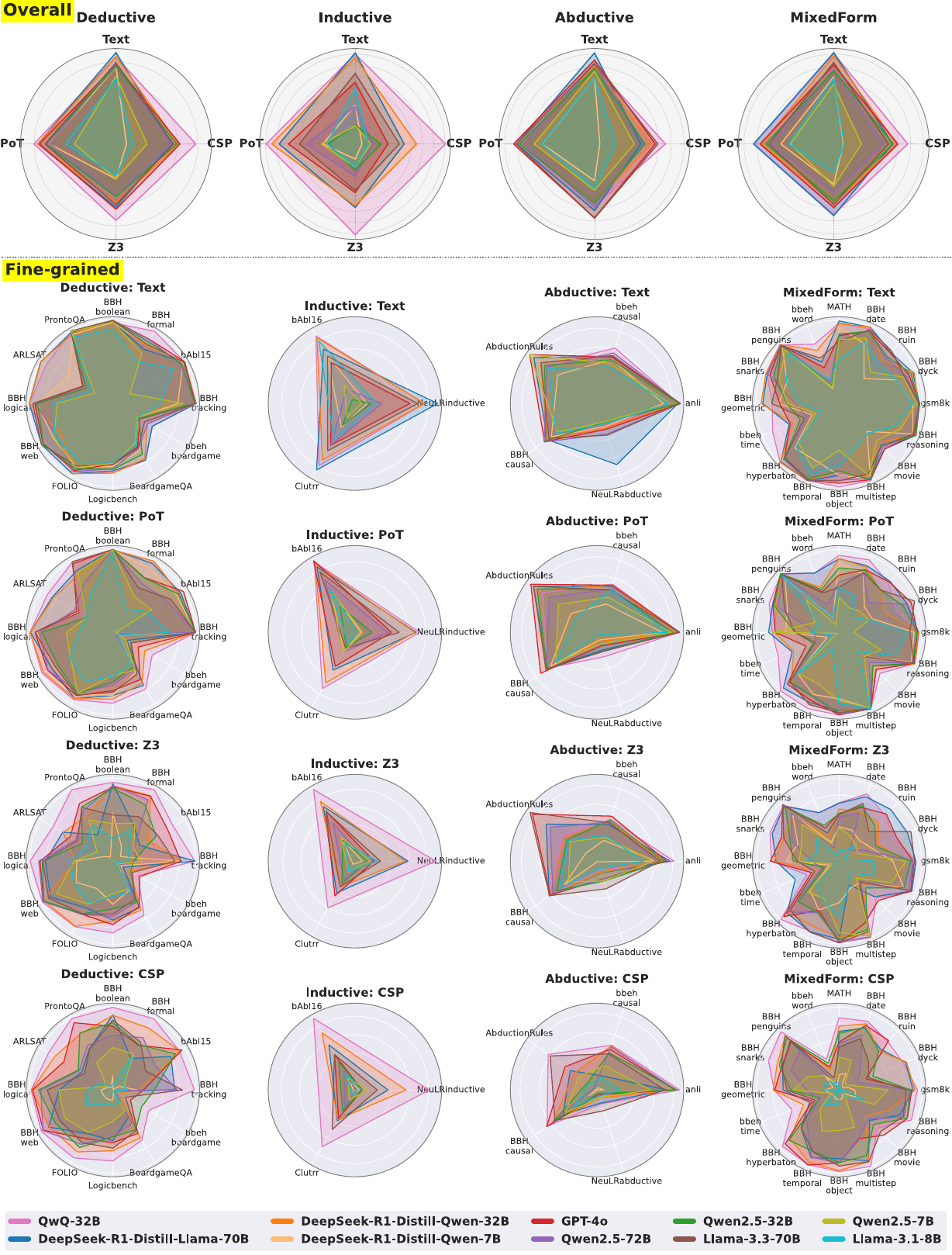}
    \caption{
    Radar plots illustrating the performance (\%) of multiple LLMs across different reasoning task types (Deductive, Inductive, Abductive, Mixed Form) and trajectory formats (Text, PoT, Z3, CSP). Overall (top 1 × 4) shows aggregated performance by reasoning type and format. Fine-grained (below 4 × 4) present fine-grained results on individual tasks
    }
    \label{fig:part1-dim1-radar_chart}
\end{figure*}

As shown in Figure~\ref{fig:part1-dim1-radar_chart}, the radar chart (Overall+ Fine-grained) illustrates the model's performance under different task types and trajectory formats. 
The complete results can be found in Appendix \ref{app:Complete results for different models}.

\paragraph{\textit{Thinking} model outperforms \textit{Instruct model}}
From the overall part, we can observe that
series of \emph{Thinking} models (e.g., QwQ-32B, etc.) outperform the \emph{Instruct} series in most tasks, especially in the Inductive and Mixed-Form tasks. 
The disparities between them reflect that the \textit{Thinking} mode can better elicit the LLM to provide reliable trajectories for formal reasoning.
Previous evaluations~\cite{xu2025large} have demonstrated a similar finding that \emph{Instruct} models have achieved unsatisfactory results in inductive reasoning, but they do not provide the suggestion that the \emph{Thinking} model can perform well.

\vspace{-0.1cm}
\paragraph{Text outperforms formal languages, except for QwQ-32B}
Most models outperform formal languages in the Text trajectory format. In the Fine-grained section, as the trajectory format shifts from Text to CSP, the radar map coverage area gradually decreases, especially in the bbeh series of subtasks. However, QwQ-32B is the only model that stays ahead in all tasks and trajectories, maintaining a high level of performance in all formal languages.

\paragraph{Formal language performance drops significantly on difficult tasks}
Models can achieve comparable or even better performance than Text with formal languages in simple tasks (e.g., Z3, CSP in Deducitve-\textit{BBH\_web}), but the performance of formal languages drops off substantially in complex tasks(e.g., Deductive-\textit{bbeh\_boardgameQA}). This phenomenon again suggests that current large models are better at using non-formal languages when expressing complex logic. Possible reasons include: 1) the model training process is dominated by natural language, with a scarcity of formal language samples; and 2) the model lacks augmentation for difficult and complex problems. The performance of text formatting is average, while formal language significantly decreases. It is worth noting that GPT-4o's performance in this area is relatively stable, possibly due to its optimization in data.

\paragraph{Small models perform poorly on formal language}
Both \textit{Instruct} and \textit{Thinking} small models have acceptable overall performance under Text, but when dealing with formal languages, the performance drops rapidly. 
Taking R1-Distill-Qwen-7B as an example, its performance under the CSP trajectory is even significantly lower than similar \textit{Instruct} models, indicating that the Thinking mechanism is difficult to effectively support formal language reasoning at low parameter scales. In addition, in high complexity tasks such as \textit{bbeh-time}, \textit{bbeh-shuffle}, etc., the small model is almost completely ineffective in structured trajectories such as Z3 and CSP, and it is difficult to complete the basic logical steps, which shows its serious lack of ability to deal with formal reasoning problems.

Overall, all models except QwQ-32B show a continuous performance degradation in the trajectory format change from Text to formal language (PoT, Z3, CSP). This phenomenon suggests that the current mainstream LLMs are more adept at handling natural language tasks, while they are still deficient in formal language reasoning.

\subsection{Different Reasoning Tasks Prefer Different Trajectory Format}
\label{sec: task_preference}

\begin{figure*}[tp]
    \centering
    \includegraphics[width=1\textwidth]{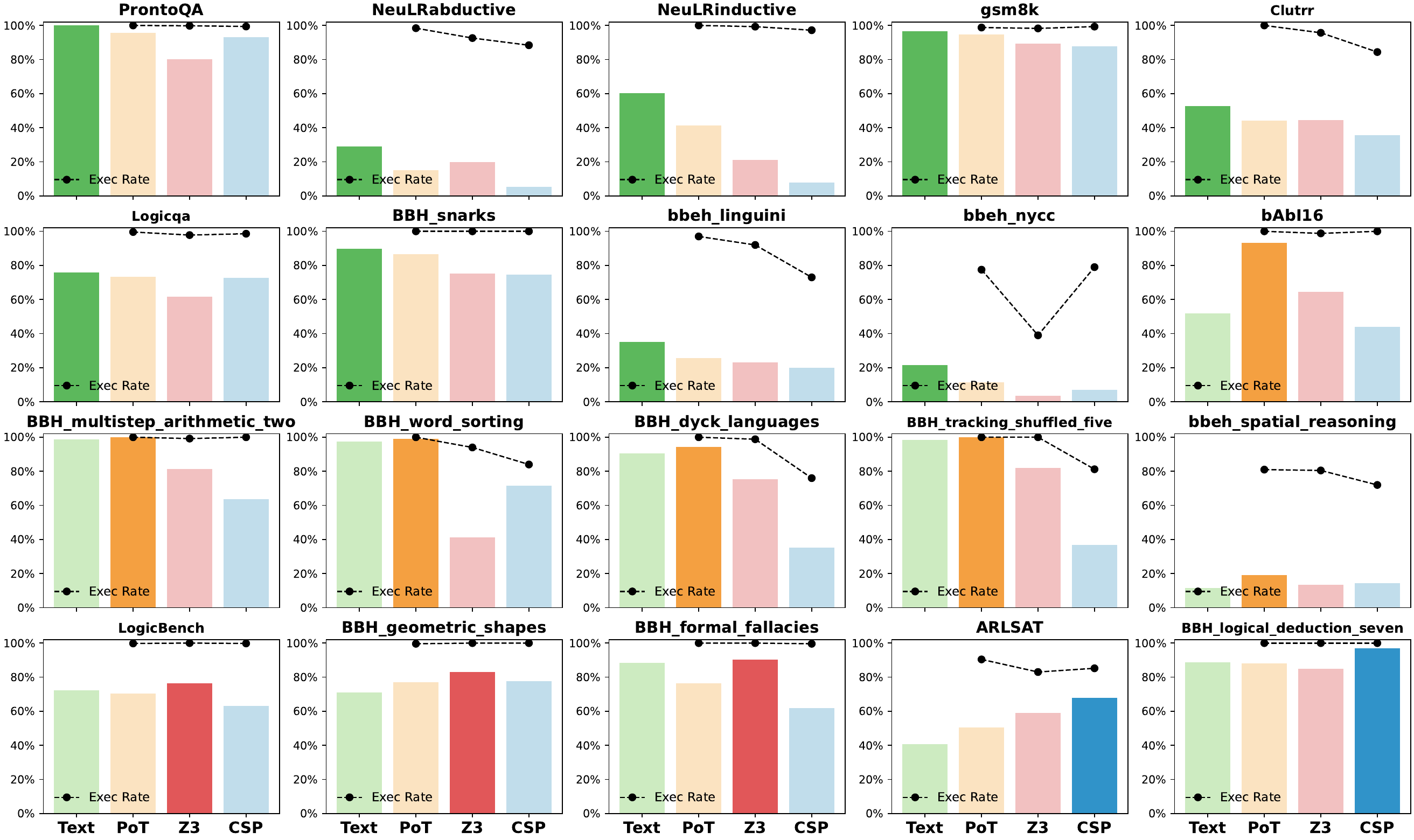}
    \caption{Preferred reasoning task performance across different trajectory formats (\textcolor[rgb]{0.3608,0.7216,0.3608}{Text}, \textcolor[rgb]{0.9569,0.6275,0.2549}{PoT}, \textcolor[rgb]{0.8824,0.3412,0.3490}{Z3}, \textcolor[rgb]{0.1882,0.5765,0.7882}{CSP})  in GPT-4o results. Each subplot shows task accuracy under different formats, with execution rate (Exec Rate) plotted as a black line. The highlighted bars represent the most preferred trajectory format for each task.}
    \label{fig:part1-2-g4o_results}
\end{figure*}

In this section, we use the GPT-4o result as an anchor point to conduct a detailed analysis of how different tasks exhibit varying preferences for trajectory formats.
As shown in Figure~\ref{fig:part1-2-g4o_results}, GPT-4o exhibits diverse preferences across trajectory formats. Below, we summarize the main observations.

\paragraph{Text performs better in language comprehension and open-ended tasks}
First, in tasks such as \textit{BBH\_snarks}, \textit{bbeh\_linguini}, \textit{bbeh\_nycc}, Text is closer to the nature of the task in humor comprehension, linguistic style recognition, and fuzzy semantic parsing, and is superior to formal language.
Secondly, in induction and abduction tasks such as \textit{AbductionRules}, \textit{NeuLRabductive}, \textit{NeuLRinductive}, and \textit{Clutrr}, where reasoning relies on linguistic expressions, the Text format is more advantageous.
In addition, \textit{LogicQA}, although categorized as a logic task, is more akin to a general knowledge quiz. It originates from the Chinese Civil Service Exam, where textual ability plays a dominant role in performance. (Cases in Figure~\ref{fig:Case for Text})

\paragraph{Well-structured tasks prefer PoT }

PoT format is particularly effective in tasks with strong structural characteristics, such as numerical computation and symbolic reasoning tasks like \textit{BBH\_dyck\_languages} and \textit{BBH\_word\_sorting}. In these settings, PoT enables efficient computation and facilitates the handling of rules involving nesting and ordering.
Additionally, in tasks that involve temporal sequences, object tracking, and spatial reasoning, such as \textit{bAbI16}, \textit{bbeh\_shuffled\_objects}, and \textit{bbeh\_spatial\_reasoning}, PoT demonstrates strong performance by leveraging programmatic trajectories to clearly express intermediate states and transformation processes. (Case in Figure~\ref{fig:Case for PoT})

\paragraph{Z3 handles formal and FOL reasoning well.}
Z3 format shows a good adaptation to formal logic tasks, especially in tasks with strict logical rules:
\textit{LogicBench}, \textit{BBH\_formal\_fallacies}, \textit{BBH\_logical\_deduction}. This type of task is essentially convertible to first-order logical expressions, so using an SMT solver (e.g., Z3) as the trajectory language is more suitable. In addition, \textit{BBH\_geometric\_shapes} involves spatial reasoning, where the boolean logical expressiveness of Z3 is more advantageous. (Case in Figure~\ref{fig:Case for CSP})

\paragraph{CSP shows advantages in complex constraints}
CSP format shows advantages in some structured logic tasks, such as \textit{BBH\_logical\_deduction}, a result consistent with the findings of Logic-LM~\cite{pan2023logic}.
More interestingly, in \textit{ARLSAT}, a task derived from the Law School Admission Test, CSP also achieves the optimal result, which contrasts with the previous~\cite{pan2023logic} literature's conclusion that Z3 is better suited for this task. 
This difference may stem from the characteristics of the tasks themselves; in ARLSAT, the stems of the questions typically contain constraints, which are more consistent in form with the way CSPs are expressed. (Case in Figure~\ref{fig:Case for CSP})

Beyond the four dimensions mentioned above, we can observe that execution success rate (Exec Rate) is also a key factor underlying the differences among various forms of language.
Moreover, gsm8k achieves its best performance under the Text format, which is inconsistent with findings from previous studies (e.g.,~\citet{ye2023satlm, he23solving}). This discrepancy may be attributed to two factors: 1) Prior work often involves task-specific optimization for mathematical reasoning; 2) Current large language models are trained on substantial amounts of mathematical natural language reasoning data, which enhances their generalization ability in Text formats.

Overall, task trajectory alignment plays a critical role. Different tasks exhibit preferences for specific trajectory formats—some tasks are inherently better suited to certain formal representations, and using inappropriate formats may even hinder model performance.
Therefore, when constructing multi-trajectory training or evaluation frameworks, it is important to carefully consider the alignment among task structure, target language, and model capabilities.

\begin{figure*}[tp]
    \centering
    \begin{subfigure}{0.49\textwidth}
        \centering
        \includegraphics[width=\linewidth]{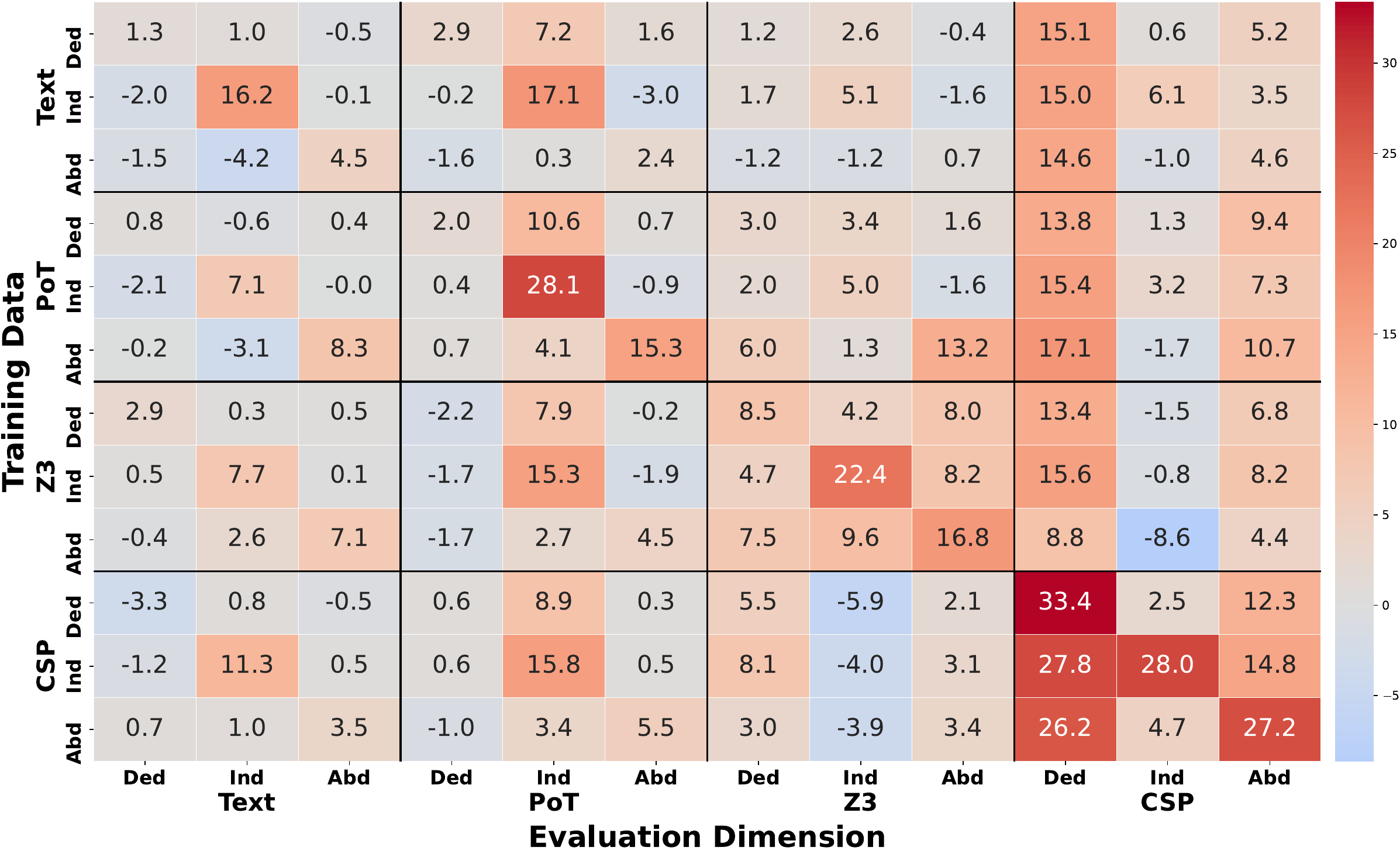}
        \caption{Fine-grained by Trajectory Format.}
        \label{fig:fine-left}
    \end{subfigure}
    \hfill
    \begin{subfigure}{0.49\textwidth}
        \centering
        \includegraphics[width=\linewidth]{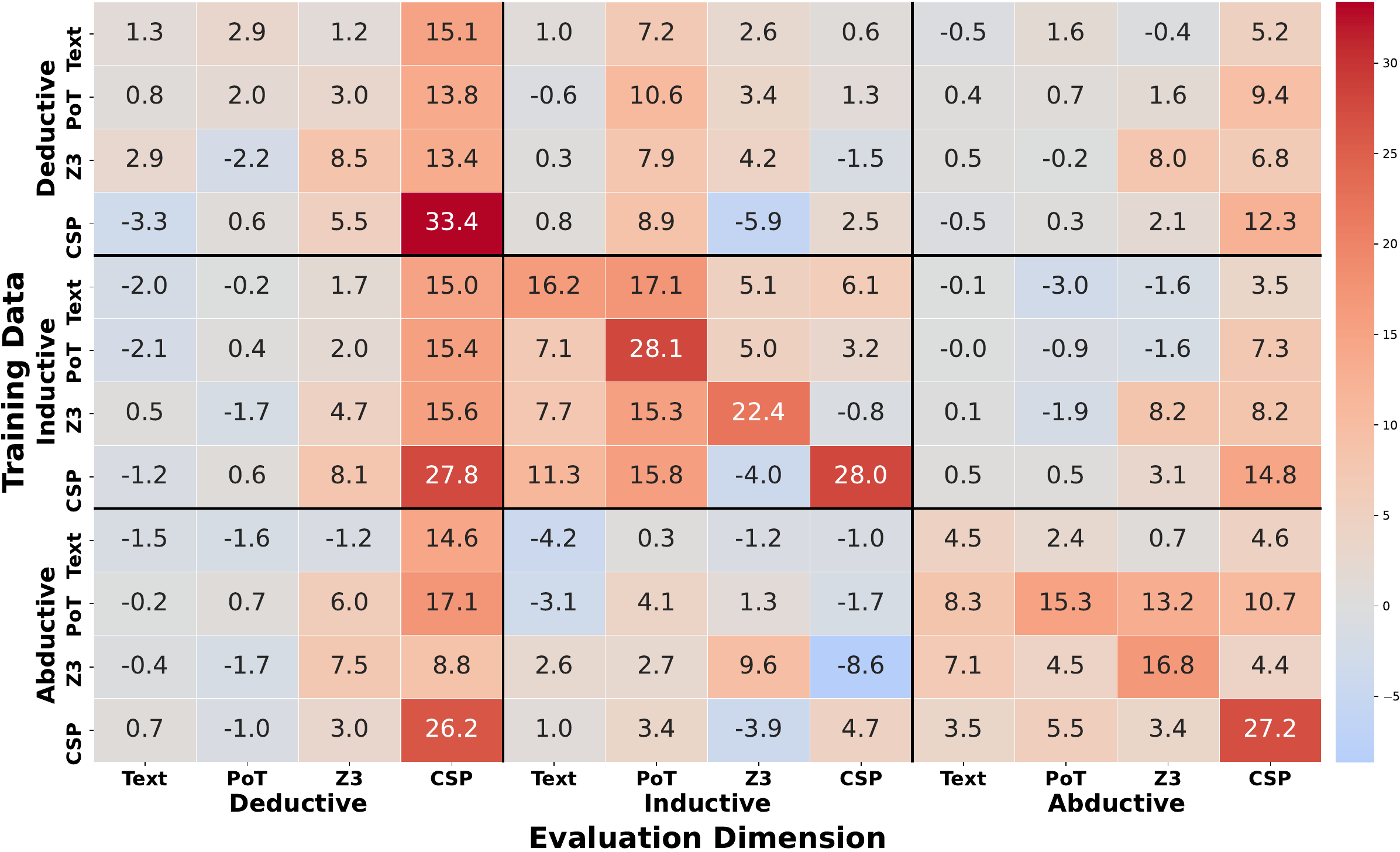}
        \caption{Fine-grained by Reasoning Type.}
        \label{fig:fine-right}
    \end{subfigure}
    \caption{Generalization performance across \textbf{fine-grained} (task type × format) configurations. Each cell shows the performance gain (\(\Delta\)) from training on the row configuration and evaluating on the column configuration}
    \label{fig:Fine}
\end{figure*}

\begin{figure}[htp]
    \centering
    \begin{subfigure}{0.49\columnwidth}
        \centering
        \includegraphics[width=\linewidth]{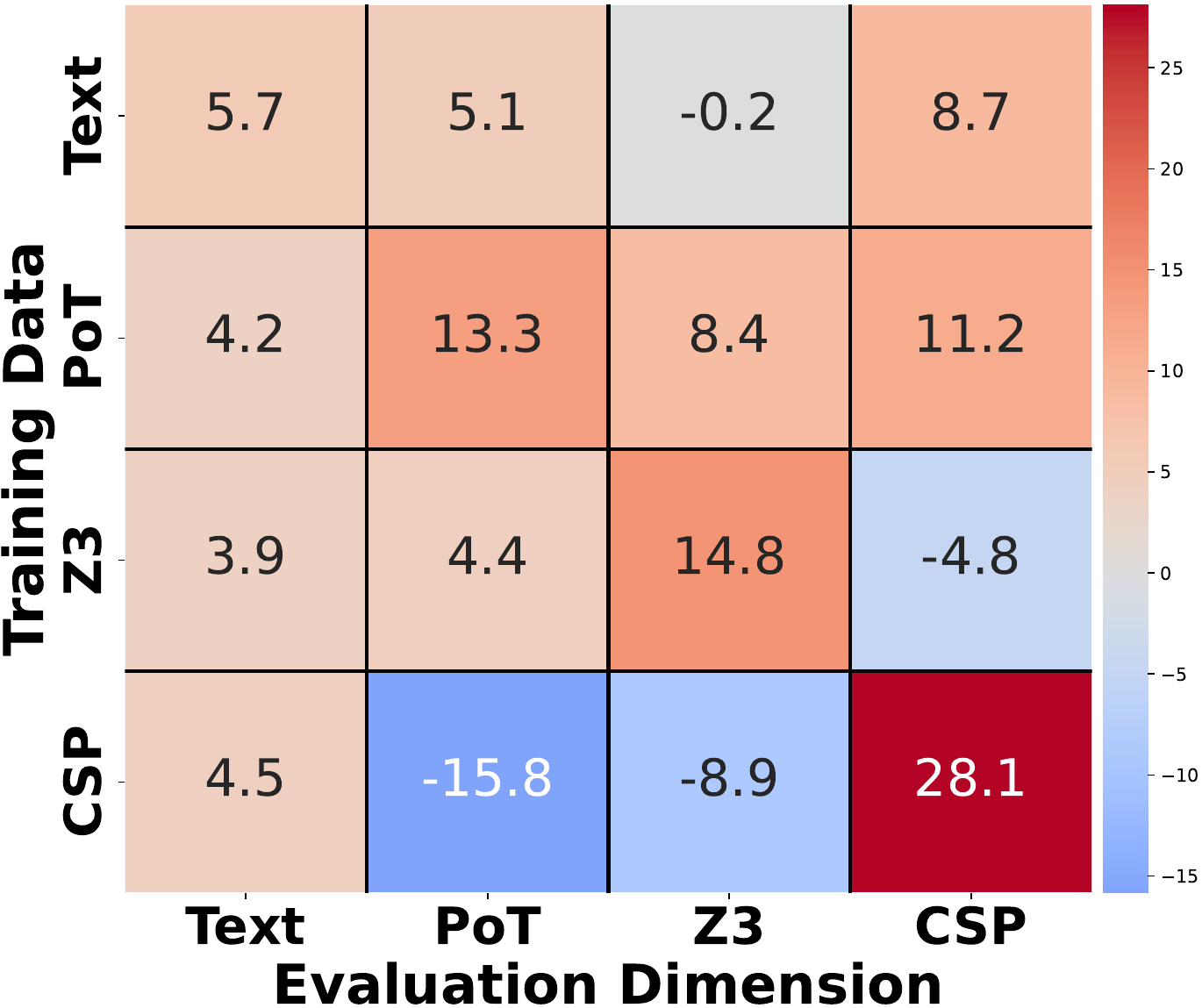}
        \caption{Trajectory Format.}
        \label{fig:Coarse_left}
    \end{subfigure}
    \hfill
    \begin{subfigure}{0.49\columnwidth}
        \centering
        \includegraphics[width=\linewidth]{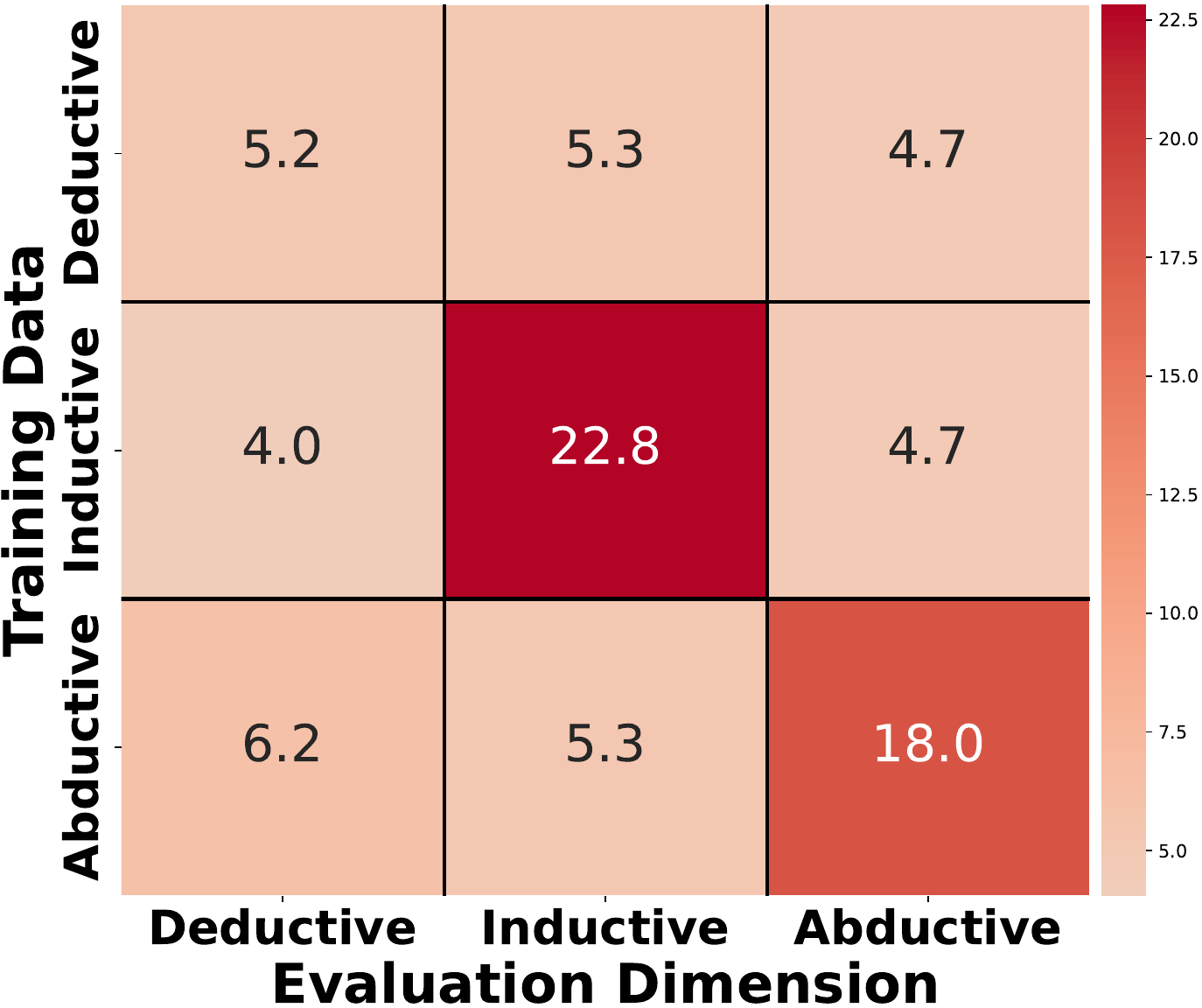}
        \caption{Reasoning Type.}
        \label{fig:Coarse_right}
    \end{subfigure}
    \caption{Generalization performance across reasoning types and trajectory formats (\textbf{coarse-grained} analysis). Each cell reports the performance gain (\(\Delta\)) when training on the row group and evaluating on the column group.}
    \label{fig:Coarse}
\end{figure}

\section{PART II: Generalization Analysis across Reasoning Tasks and Trajectory Formats}
\label{sec:part-two}

\subsection{Setup and Visualization Overview}

We collected the training split of the evaluation dataset, categorized into three reasoning types: Deductive, Inductive, and Abductive (excluding Mixed-Form due to variable control challenges). Each training instance is represented in four trajectory formats: Text, PoT, Z3, and CSP. Details are provided in Section~\ref{sec:formal_data_construct}.

Based on the above data, we conduct two sets of analytical experiments: coarse-grained and fine-grained. 1): \textbf{Coarse-grained} experiments, as shown in Figure~\ref{fig:Coarse}, involve training on 7 groups of data (3 reasoning types + 4 trajectory formats), each mixed with general-domain data, and evaluating on the same 3 reasoning types and 4 formats. 2): \textbf{Fine-grained} experiments, as shown in Figure~\ref{fig:Fine}, involve training on 12 groups of data (3 reasoning types × 4 trajectory formats), each mixed with general-domain data, and evaluating across all 12 combinations of reasoning types and formats.
Each heatmap cell shows the performance gain (\(\Delta\)) when training on the configuration in the row and evaluating on the configuration in the column.
The performance gain reflects the improvement introduced by our constructed data when mixed with the general-domain data (Trained on Qwen-2.5-7B).

\begin{table*}[htbp]
  \centering
  \resizebox{\linewidth}{!}{%
    \begin{tabular}{cccccccccc}
    \toprule
    \multirow{2}[4]{*}{\textbf{Model}} & \textbf{Text} & \multicolumn{2}{c}{\textbf{PoT}} & \multicolumn{2}{c}{\textbf{Z3}} & \multicolumn{2}{c}{\textbf{CSP}} & \multicolumn{2}{c}{\textbf{Avg}} \\
\cmidrule{2-10}          & \textbf{Acc} & \textbf{Acc} & \textbf{Exec Rate} & \textbf{Acc} & \textbf{Exec Rate} & \textbf{Acc} & \textbf{Exec Rate} & \textbf{Acc} & \textbf{Exec Rate} \\
    \midrule
    \textbf{GPT-4o} & 66.7 & 64.0  & 91.5 & 54.5 & 87.4 & 53.0  & 83.98 & 59.0  & 87.6 \\
    \textbf{Qwen2.5-7B-Instruct} 
    & \underline{52.3} 
    & \underline{37.0}  
    & \underline{78.6} 
    & \underline{33.0}  
    & \underline{70.0} 
    & \underline{25.0}  
    & \underline{52.1} 
    & \underline{37.0}  
    & \underline{66.9} \\
    \textbf{Qwen2.5-72B-Instruct} & 63.4 & 54.0  & 85.1 & 42.5 & 79.6 & 43.0  & 75.2 & 51.0  & 80.0 \\
    \midrule
    \midrule
    \textbf{Qwen2.5-7B-Baseline} & 49.7 & 40.0   & 75.4 & 27.1 & 68.2 & 20.0   & 52.2 & 34.0  & 65.3 \\
    \textbf{Qwen2.5-7B-Base w.Formal} 
    & 52.7\textsuperscript{\textcolor{deepgreen}{+3.0}} 
    & 44.0\textsuperscript{\textcolor{deepgreen}{+4.0}}  
    & 83.5\textsuperscript{\textcolor{deepgreen}{+8.1}} 
    & 34.8\textsuperscript{\textcolor{deepgreen}{+7.7}} 
    & 76.5\textsuperscript{\textcolor{deepgreen}{+8.3}} 
    & 37.0\textsuperscript{\textcolor{deepgreen}{+17.0}} 
    & 68.1\textsuperscript{\textcolor{deepgreen}{+15.9}} 
    & 42.0\textsuperscript{\textcolor{deepgreen}{+8.0}}  
    & 76.0\textsuperscript{\textcolor{deepgreen}{+10.7}} \\
    \bottomrule
    \end{tabular}}
  \caption{Performance of LLM on different trajectory formats before and after formal data enhancement. Accuracy (Acc) and execution rate (Exec Rate) are reported for text, PoT, Z3, and CSP formats. \textbf{Qwen2.5-7B-Baseline} denotes the baseline model trained with general data only; \textbf{Qwen2.5-7B-Base w.Formal} denotes the augmented model trained with a mixture of formal language data. Improvements after augmentation are shown in green.}
  \label{tab:part3_main_result}
\end{table*}

\subsection{Coarse-Grained Generalization Analysis}
\label{sec:coarse-generalization}

\paragraph{Significant in-domain improvement}
The strongest performance gains are observed along the diagonal, indicating that the model benefits most when the training and evaluation data come from the same group. Notably, the improvements for CSP (Train) → CSP (Eval) and Inductive (Train) → Inductive (Eval) reach 28.1 and 22.8, respectively. Combined with observations from Part-I, this can be partially attributed to the relatively low baseline performance of the Qwen2.5-7B model on the CSP and Inductive dimensions, meaning that even a small amount of in-domain data leads to significant improvement.

\paragraph{PoT transfers well, while CSP transfers poorly}
Outside the diagonal, in figure \ref{fig:Coarse_left}, PoT migrates well in Text, Z3, and CSP. This might be related to the fact that there is a lot of code data in the pre-training data. CSP, on the other hand, has an effect only on Text and CSP, with significant negative effects on PoT (-15.8) and Z3 (-8.9). This suggests that there may be structural differences among formal languages.

\paragraph{Reasoning types: all exhibit positive transfer}
The overall transfer effect is relatively balanced between the different reasoning types (Fig.\ref{fig:Coarse_right}). The relatively small improvement on Deductive itself may be related to the higher base level of the model on Deductive.

\subsection{Fine-Grained Generalization Analysis}
\label{sec:fine-grained-generalization}

\paragraph{Deductive-CSP is most easily generalized}
In Figure~\ref{fig:Fine}, all entries in the Deductive-CSP column show improvements. The inclusion of any data contributes positively to its performance. This is mainly because CSP has a relatively low baseline, and the Deductive category contains some relatively simple tasks (\textit{BBH\_logical\_deduction\_three} from 40 \% to 92\%). As a result, adding similar data leads to performance gains.

\paragraph{CSP and Z3 transfer well across reasoning types}
In Figure~\ref{fig:fine-left}, all entries (Ded/Ind/Abd) within the CSP and Z3 blocks show positive gains, indicating that regardless of reasoning type, CSP and Z3 formats can be effectively transferred.

\paragraph{Abductive transfers well across trajectory formats}
In Figure~\ref{fig:fine-right}, all entries (Text/PoT/CSP/Z3) within the Abductive block show improvements, suggesting that regardless of trajectory format, Abductive reasoning can be effectively transferred and improved.

\section{PART III: Enhancing LLMs with Formal Data}

\subsection{Formal Data Construction via RFT}
\label{sec:formal_data_construct}

To enhance model capability in formal languages, we collect the portions of current evaluation datasets that overlap with training data as part of our training set. All dataset details are provided in Table~\ref{tab:all_dataset}.
Similarly, the training data is categorized into three types: Deductive, Inductive, and Abductive, and four trajectory formats: Text, PoT, Z3, and CSP.

First, we extract up to 3,000 samples from all training data. Then, GPT-4o was chosen as the output for teacher model construction. In order to obtain high quality response data, we used \textbf{Rejection sampling Fine-Tuning (RFT)}. We used GPT-4o to sample the questions several times and then filtered out those samples whose code was executable and whose final answers are verified to be correct. The statistics of the filtered data are shown in Table~\ref{tab:entire_eval_exp_stats}. The number in parentheses after each model name indicates the amount of added data.

\subsection{Main Result}

As shown in Table~\ref{tab:part3_main_result}, the enhanced model improves accuracy by 3.0\% on Text, 7.7\% on Z3 (with an 8.3\% gain in execution rate), and 17.0\% on CSP (from 20.0\% to 37.0\%, with a 15.9\% increase in execution rate). Overall, average accuracy rises from 34.0\% to 42.0\%, and execution rate from 65.3\% to 76.0\%.

Beyond outperforming the baseline, our formal-data-enhanced model also surpasses the open-source model Qwen2.5-7B-Instruct across all formats.
Qwen2.5-7B-Base w.Formal has a smaller parameter size than Qwen2.5-72B, but the performance gap is narrowed by formal data fine-tuning. 
This suggests that formal data augmentation can effectively improve the competitiveness of small models in formal reasoning tasks.

\section{Conclusion}

In this paper, we provide a comprehensive evaluation of LLMs utilizing various formal languages to solve different categories of logical reasoning tasks.
We first develop a systematic evaluation architecture and decompose it into three dimensions.
Then, we perform a thorough evaluation across these three dimensions to show whether the current LLMs can excel in formal language utilization.
Furthermore, we explore the generalization across multiple formal languages and provide a simple but effective method on the capability enhancement for small language models.

For future directions, on the one hand, we should strive to enhance the model's reasoning capabilities in a balanced manner across different trajectory formats and task types, especially for Instruct models. At the same time, it may be valuable to construct formal language reasoning datasets in a "thinking" style.  
On the other hand, we can leverage the task-specific preferences for trajectory formats to further expand the capability boundaries of the model. One approach is to incorporate reasoning results from different trajectory formats as individual voters in a majority voting scheme. Another approach is to introduce multiple symbolic solvers for different reasoning trajectories during the thinking stage of the think model.

\newpage
\section*{Limitations}

This work provides a step toward evaluating and enhancing LLMs through formal reasoning formats, but several limitations remain.
First, the landscape of LLMs is evolving rapidly. Our experiments focus on a limited set of models available at the time, and newer models may change performance trends.
Second, while we include various reasoning types and benchmark datasets, the overall dataset coverage is limited. Our formal data augmentation is applied to a subset of tasks and may not generalize to other domains.
Third, we focus on three formal formats, "PoT, Z3, and CSP," due to their executability and popularity. However, this excludes other symbolic systems such as Lean, Prolog, Coq, or SMT-LIB, which future work could explore.
Finally, our formal data construction is based on the \textit{Instruct} model (GPT-4o). With the rise of stronger \textit{Thinking} models, generating think-style formal data may become more feasible and diverse in the future.

\bibliography{acl_latex}

\newpage
\appendix

\section{Related Work}
\label{app:related_work}
\subsection{Symbolic Solver Enhances LLM Reasoning}
The integration of symbolic solvers with large language models (LLMs) has emerged as a promising approach to enhance logical reasoning. Early efforts focused on translating natural language to first-order logic (FOL), exemplified by the creation of the MALLS dataset and the LogicLLaMA model, which demonstrated improved NL-to-FOL translation \cite{yang2023harnessing}. The Logic-LM framework further explored this direction by employing different formal languages and solvers tailored to specific reasoning tasks, such as FOL with Prover9, CSP solvers for constraint satisfaction, and Z3 for SMT problems \cite{pan2023logic}. SATLM introduced declarative prompting to generate task specifications in logical formulas for SAT solvers \cite{ye2023satlm}, while LINC utilized LLMs for semantic parsing into FOL, offloading inference to theorem provers \cite{olausson2023linc}. Subsequent research investigated strategies for improving NL-to-FOL translation through data generation and fine-tuning \cite{xiong2024strategies}, multi-step refinement of symbolic formulations \cite{wang2024logiclm++}, and the impact of pre-training data, including programming languages, on logical inference \cite{uchiyama2024programming}. Frameworks like VERUS-LM aimed for versatility by supporting various reasoning tasks with a clear separation of knowledge and queries \cite{callewaert2025verus}.

\subsection{Complex Logical Reasoning Tasks}
Evaluating the logical reasoning capabilities of LLMs necessitates challenging and diverse datasets that probe various aspects of inference. FOLIO, annotated with first-order logic, focuses on complex logical reasoning in natural language \cite{han2024folio}. ProntoQA utilizes logic programming and emphasizes chain-of-thought reasoning \cite{saparov2022language}, while LogicBench covers propositional, first-order, and non-monotonic logic with a focus on single inference rules \cite{parmar2024logicbench}. BOARDGAMEQA assesses reasoning with contradictory information and preferences \cite{kazemi2023boardgameqa}, and AR-LSAT tests analytical reasoning skills using logic constraints \cite{zhong2022analytical}. The BIG-Bench Hard (BBH) benchmark includes a wide array of challenging tasks like Boolean Expressions \cite{suzgun2022challenging}, formal fallacies \cite{suzgun2022challenging}, logical deduction \cite{suzgun2022challenging}, shuffled objects \cite{suzgun2022challenging}, and web of lies \cite{suzgun2022challenging}. Other datasets like bAbI \cite{weston2015towards}, CLUTRR \cite{sinha2019clutrr}, $\alpha$-NLI \cite{zhao2021adversarial}, AbductiveRules \cite{bhagavatula2020abductionrules}, LogiQA \cite{liu2020logiqa}, and gsm8k \cite{cobbe2021training} target specific reasoning types such as deductive, inductive, abductive, temporal, spatial, and mathematical reasoning. The variety in these datasets and their annotations highlights the multifaceted nature of complex reasoning and the ongoing efforts to evaluate and enhance LLMs in this domain.

\section{Details of Datasets}
\label{app:datasets}

Table~\ref{tab:all_dataset} provides a comprehensive overview of all datasets used in our study. Each dataset is annotated with its reasoning type (Deductive, Inductive, Abductive, or Mixed-Form), along with the number of evaluation and training examples. We also include the original source for each dataset.

The classification follows our taxonomy introduced in Section~\ref{sec:preliminary-one}. In particular:
\begin{itemize}
  \item \textbf{Deductive} datasets include tasks that require formal logical reasoning based on explicit rules or premises.
  \item \textbf{Inductive} datasets focus on pattern discovery and generalization from limited examples.
  \item \textbf{Abductive} datasets involve generating plausible explanations under uncertainty.
  \item \textbf{Mixed-Form} includes tasks with hybrid or ambiguous reasoning types, further grouped into subcategories such as Temporal, NLU, Symbolic, Spatial, Knowledge, and Math.
\end{itemize}

Some datasets (e.g., BBH and bbeh) are split into finer task categories, each treated independently during evaluation. For large-scale datasets like GSM8K and MATH, we use a subset of examples (denoted by *) to maintain balance across task types.

This dataset collection forms the foundation for our evaluation across models, trajectory formats, and reasoning types.
\begin{table*}[htbp]
  \centering
    \resizebox{!}{12cm}{%
    \begin{tabular}{ccccc}
    \toprule
    \textbf{Type} & \textbf{Dataset} & \textbf{Eval} & \textbf{Train} & \textbf{Original Source} \\
    \midrule
    \multirow{21}[2]{*}{\textbf{Deductive}} 
    & FOLIO & 134   & 674   & \citet{han2024folio} \\
    & ProntoQA & 500   & 50818 & \citet{han2024folio} \\
    & LogicBench & 500   & 12908 & \citet{parmar2024logicbench} \\
    & BOARDGAMEQA & 14K   & 750K  & \citet{kazemi2023boardgameqa} \\
    & AR-LSAT & 230   & 1629  & \citet{zhong2021ar} \\
    & BBH (Boolean Expression) & 250   & -     & \citet{suzgun2022challenging} \\
    & bbeh (Boolean Expressions) & 200   & -     & \citet{kazemi2025big} \\
    & BBH (formal\_fallacies) & 250   & -     & \citet{suzgun2022challenging} \\
    & bbeh (Zebra Puzzles) & 200   & -     & \citet{kazemi2025big} \\
    & BBH (logical\_deductive\_five\_objects) & 250   & -     & \citet{suzgun2022challenging} \\
    & BBH (logical\_deductive\_seven\_objects) & 250   & -     & \citet{suzgun2022challenging} \\
    & BBH (logical\_deductive\_three\_objects) & 250   & -     & \citet{suzgun2022challenging} \\
    & bbeh (Boardgame QA) & 200   & -     & \citet{kazemi2025big} \\
    & BBH (tracking\_shuffled\_objects\_five\_objects) & 250   & -     & \citet{suzgun2022challenging} \\
    & BBH (tracking\_shuffled\_objects\_seven\_objects) & 250   & -     & \citet{suzgun2022challenging} \\
    & BBH (tracking\_shuffled\_objects\_three\_objects) & 250   & -     & \citet{suzgun2022challenging} \\
    & bbeh (Shuffled Objects) & 200   & -     & \citet{kazemi2025big} \\
    & BBH (web\_of\_lies) & 250   & -     & \citet{suzgun2022challenging} \\
    & bbeh (Web of Lies) & 200   & -     & \citet{kazemi2025big} \\
    & bAbI-15 & 1000  & 900   & \citet{weston2016towards} \\
    & NeuLR-deductive & 7001  & -     & \citet{xu2025large} \\
    \midrule
    \multirow{3}[2]{*}{\textbf{Inductive}} 
    & CLUTRR & 1042  & 2452  & \citet{sinha2019clutrr} \\
    & bAbI-16 & 1000  & 900   & \citet{weston2016towards} \\
    & NeuLR-inductive & 7001  & -     & \citet{xu2025large} \\
    \midrule
    \multirow{5}[2]{*}{\textbf{Abductive}} 
    & $\alpha$-NLI & 3059  & 169k  & \citet{valentino2022case} \\
    & AbductiveRules & 2536  & 8848  & \citet{young2022abductionrules} \\
    & BBH (causal\_judgement) & 250   & -     & \citet{suzgun2022challenging} \\
    & bbeh (Causal Understanding) & 200   & -     & \citet{kazemi2025big} \\
    & NeuLR-abductive & 6001  & -     & \citet{xu2025large} \\
    \midrule
    \textbf{Mixed-Form}
    &   &    &      &   \\
    \cdashline{1-5}
    Logical 
    & LogiQA & 1572  & -     & \citet{liu2021logiqa} \\
    \cdashline{1-5}
    Temporal 
    & BBH (date\_understanding) & 250   & -     & \citet{suzgun2022challenging} \\
    & bbeh (Time Arithmetic) & 200   & -     & \citet{kazemi2025big} \\
    & BBH (temporal\_sequences) & 250   & -     & \citet{suzgun2022challenging} \\
    & bbeh (Temporal Sequences) & 200   & -     & \citet{kazemi2025big} \\
    \cdashline{1-5}
    NLU
    & BBH (disambiguation\_qa) & 250   & -     & \citet{suzgun2022challenging} \\
    & bbeh (Disambiguation QA) & 200   & -     & \citet{kazemi2025big} \\
    & BBH (hyperbaton) & 250   & -     & \citet{suzgun2022challenging} \\
    & bbeh (Hyperbaton) & 200   & -     & \citet{kazemi2025big} \\
    & BBH (ruin\_names) & 250   & -     & \citet{suzgun2022challenging} \\
    & bbeh (New Yorker Cartoon Caption) & 200   & -     & \citet{kazemi2025big} \\
    & BBH (salient\_translation\_error\_detection) & 250   & -     & \citet{suzgun2022challenging} \\
    & bbeh (Linguini) & 200   & -     & \citet{kazemi2025big} \\
    & BBH (snarks) & 250   & -     & \citet{suzgun2022challenging} \\
    & bbeh (SARC Triples) & 200   & -     & \citet{kazemi2025big} \\
    \cdashline{1-5}
     Symbolic 
    & BBH (dyck\_languages) & 250   & -     & \citet{suzgun2022challenging} \\
    & bbeh (Dyck Language) & 200   & -     & \citet{kazemi2025big} \\
    & BBH (word\_sorting) & 250   & -     & \citet{suzgun2022challenging} \\
    & bbeh (Word Sorting) & 200   & -     & \citet{kazemi2025big} \\
    \cdashline{1-5}
    Space 
    & BBH (geometric\_shapes) & 250   & -     & \citet{suzgun2022challenging} \\
    & bbeh (Geometric Shapes) & 200   & -     & \citet{kazemi2025big} \\
    & BBH (navigate) & 250   & -     & \citet{suzgun2022challenging} \\
    & bbeh (Spatial Reasoning) & 200   & -     & \citet{kazemi2025big} \\
    \cdashline{1-5}
    Table
    & BBH (penguins\_in\_a\_table) & 250   & -     & \citet{suzgun2022challenging} \\
    & bbeh (Buggy Tables) & 200   & -     & \citet{kazemi2025big} \\
    \cdashline{1-5}
     Knowledge 
    & BBH (moive\_recommendation) & 250   & -     & \citet{suzgun2022challenging} \\
    & bbeh (Movie Recommendation) & 200   & -     & \citet{kazemi2025big} \\
    & BBH (sports\_understanding) & 250   & -     & \citet{suzgun2022challenging} \\
    & bbeh (SportQA) & 200   & -     & \citet{kazemi2025big} \\
    \cdashline{1-5}
    MATH
    & GSM8K & 1319  & *8790     & \citet{cobbe2021training} \\
    & MATH  & 5000  & *7500     & \citet{hendrycks2measuring} \\
    & BBH (multistep\_arithmetaic\_two) & 250   & -     & \citet{suzgun2022challenging} \\
    & bbeh (Multi-step Arithmetic) & 200   & -     & \citet{kazemi2025big} \\
    & BBH (object\_counting) & 250   & -     & \citet{suzgun2022challenging} \\
    & bbeh (Object Counting) & 200   & -     & \citet{kazemi2025big} \\
    & BBH (reasoning\_about\_colored\_objects) & 250   & -     & \citet{suzgun2022challenging} \\
    & bbeh (Object Properties) & 250   & -     & \citet{suzgun2022challenging} \\
    \bottomrule
    \end{tabular}%
    }
    \caption{Complex Logical Reasoning data categorization, data statistics, and sources.}
  \label{tab:all_dataset}%
\end{table*}%

\section{Detail of Trajectory Format}
\label{app:formal-trajectories}

We extend the unified trajectory formulation to three specific formal languages: Python (PoT), Z3, and CSP. Each trajectory consists of a sequence of symbolic steps, which are executed by an external engine to compute the final answer.

We denote the model-generated trajectory as:
\begin{align}
\mathcal{LLM}(Q) = \langle s_1, s_2, \dots, s_n \rangle 
\quad \xrightarrow{\text{Exec}} \quad A
\end{align}
Where $Q$ is the input query, each $s_i$ is a step in a domain-specific language, and $A$ is the final answer produced by executing the trajectory.

\subsection*{Python (PoT) Trajectory}

In the Python format, each step $s_i$ is a syntactically valid Python statement. The trajectory consists of variable assignments, arithmetic operations, control logic, and ends with a \texttt{print(A)} statement.

The Python trajectory is formalized as:
\begin{align}
&\mathcal{LLM}_{\text{Python}}(Q) = 
\\&\quad  
\langle 
\texttt{stmt}_1, \texttt{stmt}_2, \dots, \texttt{stmt}_n,  
\\
&\quad\texttt{print}(A) \rangle 
\quad \xrightarrow{\text{Python 3.12}} \quad A
\end{align}

This trajectory is interpreted and executed sequentially using a Python 3.12 interpreter. 

\subsection*{Z3 Trajectory}

Inspired by Logic-LM~\citep{pan2023logic}, for Z3, the reasoning trajectory is constructed using the Z3 theorem prover. A typical trajectory includes symbolic variable declarations such as \texttt{x = Int('x')}, followed by logical assertions like \texttt{s.add(x > 1, x < 5)}, and ends with solver calls \texttt{s.check()} and \texttt{s.model()} to extract a result.

We represent the Z3 trajectory as:
\begin{align}
&\mathcal{LLM}_{\text{Z3}}(Q) = \\
&\quad \langle 
\texttt{Declare},\ \texttt{Assert}_1, \dots, \texttt{Assert}_k, 
\\
&\quad\texttt{CheckSat},
\texttt{print(A)} \rangle 
\quad \xrightarrow{\text{Z3 Solver}} \quad A
\end{align}

Z3 supports a wide range of built-in logical theories, such as integer arithmetic, arrays, and bit-vectors.

\subsection*{CSP Trajectory}

Constraint Satisfaction Problems (CSPs) are defined by a triple $(X, D, C)$, where 
$X=\{x_1, \dots, x_n\}$ denotes variables, 
$D = \{D_1, \dots, D_n\}$ their domains, and 
$C = \{C_1, \dots, C_m\}$ the set of constraints. 
Each constraint $C_j = \langle t_j, R_j \rangle$ is defined over a subset of variables and a relation on their domains.

The CSP trajectory is modeled as:
\begin{align}
&\mathcal{LLM}_{\text{CSP}}(Q) 
  =\\ &\quad\langle\, \texttt{AddVar}_1, \dots, \texttt{AddVar}_n, \notag \texttt{AddConst}_1, \dots, \\
  &\quad\texttt{AddConst}_m, \notag \texttt{GetSolution}, \notag \\&\quad\texttt{print(A)} \rangle 
    \quad \xrightarrow{\text{python-constraint}} \quad A
\end{align}

The execution uses the \texttt{python-constraint} solver. Variables are added through \texttt{addVariable()}, constraints through \texttt{addConstraint()}, and solutions are obtained via \texttt{getSolution()} or \texttt{getSolutions()}. The solver applies standard algorithms such as backtracking and constraint propagation. 

While Prover9-FOL supports classical first-order logic, we choose Z3 for its broader practical applicability. Z3 not only supports FOL reasoning but also natively handles richer theories such as integers, arrays, and linear arithmetic. This allows it to express a wider range of constraints found in real-world reasoning tasks.

\section{Implementation Setups}
\label{app:implementation_setups}

\subsection{Evaluations Details}
\label{app:eval_setups}
In the inference phase, we use the vLLM \citep{kwon2023efficient} framework for deployment. The inference configuration adopts greedy decoding strategy and sets the maximum generation length to 16K tokens.
For the evaluation of model output, we adopt Qwen-2.5-72B-Instruct as the model evaluator to score.

\subsection{Training Details}
In terms of training implementation, we use Megatron-LM as the training framework with the following configurations: a cosine learning rate schedule is adopted with an initial learning rate of 1e-5, a warmup ratio of 0.03, and the learning rate decays to 0; the maximum sequence length is set to 8192, with a global batch size of 128, and the number of training epochs is set to 3. All experiments are completed with Supervised Fine-tuning (SFT) on a computing cluster consisting of 32 NVIDIA A100 GPUs.

\section{Complete results for different models}
\label{app:Complete results for different models}
\begin{table*}[htbp]
  \centering
    \begin{tabular}{ccc}
    \toprule
    Section & Number & Model \\
    \midrule
    \multirow{8}[2]{*}{PART-I} & \multirow{8}[2]{*}{\makecell{4 Thinking-Model\\+\\6 Instruct-Model\\=10}} & \textbf{QwQ-32B} \\
          &       & DeepSeek-R1-Distill-Llama-70B \\
          &       & DeepSeek-R1-Distill-Qwen-32B \\
          &       & DeepSeek-R1-Distill-Qwen-7B \\
          &       & \textbf{GPT-4o} \\
          &       & Qwen2.5-72B \\
          &       & Qwen2.5-32B \\
          &       & Llama-3.3-70B \\
          &       & \textbf{Qwen2.5-7B} \\
          &       & Llama-3.1-8B \\
    \midrule
    \multirow{19}[4]{*}{PART-II} & \multirow{7}[2]{*}{\makecell{3 (Deductive, Inductive, Abductive)\\+\\4 (Text, PoT, Z3, CSP)\\=7}} & Qwen2.5-7B-Base.w. Deductive (+5653) \\
          &       & Qwen2.5-7B-Base.w. Inductive (+4947) \\
          &       & Qwen2.5-7B-Base.w. Abductive (+6557) \\
          &       & Qwen2.5-7B-Base.w. Text (+7384)\\
          &       & Qwen2.5-7B-Base.w. PoT (+7448)\\
          &       & Qwen2.5-7B-Base.w. Z3 (+6882) \\
          &       & Qwen2.5-7B-Base.w. CSP (+6346)\\
\cmidrule{2-3}          & \multirow{12}[2]{*}{\makecell{3 (Deductive, Inductive, Abductive)\\×\\4 (Text, PoT, Z3, CSP)\\=12}} & Qwen2.5-7B-Base.w. Deductive\_Text (1376) \\
          &       & Qwen2.5-7B-Base.w. Deductive\_PoT (+1393) \\
          &       & Qwen2.5-7B-Base.w. Deductive\_Z3 (+1374) \\
          &       & Qwen2.5-7B-Base.w. Deductive\_CSP (+1510)\\
          &       & Qwen2.5-7B-Base.w. Inductive\_Text (+1263)\\
          &       & Qwen2.5-7B-Base.w. Inductive\_PoT (+1476)\\
          &       & Qwen2.5-7B-Base.w. Inductive\_Z3 (+1166)\\
          &       & Qwen2.5-7B-Base.w. Inductive\_CSP (+1042)\\
          &       & Qwen2.5-7B-Base.w. Abductive\_Text (+1820)\\
          &       & Qwen2.5-7B-Base.w. Abductive\_PoT (+1775)\\
          &       & Qwen2.5-7B-Base.w. Abductive\_Z3 (+1667)\\
          &       & Qwen2.5-7B-Base.w. Abductive\_CSP (+1295)\\
    \midrule
    \multirow{2}[2]{*}{PART-III} & \multirow{2}[2]{*}{\makecell{1 Baseline-Model+\\1 Formal Data Enhanced Model =2}} & Qwen2.5-7B-Baseline* (15k) \\
          &       & \textbf{Qwen2.5-7B-Base.w. Formal (+28060)} \\
    \midrule
    ALL   & 31    & - \\
    \bottomrule
    \end{tabular}%
  \caption{Comprehensive Overview of Model Evaluation Experiments in the Entire Paper (Models in Bold Are Presented with Full Results Later).  Parentheses after the model in PART-II indicate the corresponding amount of data. All data are based on the 15k generic data of Qwen2.5-7B-Baseline*, plus (+) the corresponding amount of our synthetic data.}

  \label{tab:entire_eval_exp_stats}%
\end{table*}

As shown in Table~\ref{tab:entire_eval_exp_stats}, we evaluated a total of 31 models across the three parts of this paper. Due to space constraints, we present the results of several representative models here: QwQ-32B (Table~\ref{tab:QwQ-32B}), GPT-4o (Table~\ref{tab:GPT-4o}), Qwen2.5-7B (Table~\ref{tab:Qwen2.5-7B}), and Qwen2.5-7B-Base w. Formal (Table~\ref{tab:Qwen2.5-7B-Base.w.Formal}). The complete results are provided in the supplementary files in Excel format.

\section{Case Study}
\label{app:case_study}

\begin{figure*}[tp]
    \centering
    \includegraphics[width=0.9\textwidth]{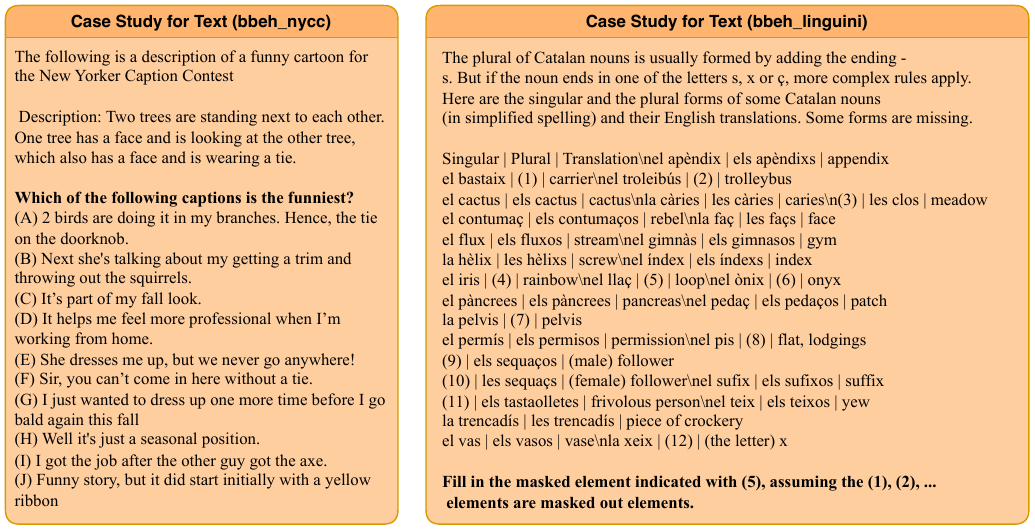}
    \caption{Case for Text. \ref{sec: task_preference}. These issues are closely related to semantic understanding. They are better suited for textual representation and are difficult to address using formalized languages. Return to section \ref{sec: task_preference}. } 
    \label{fig:Case for Text}
\end{figure*}

\begin{figure*}[tp]
    \centering
    \includegraphics[width=0.9\textwidth]{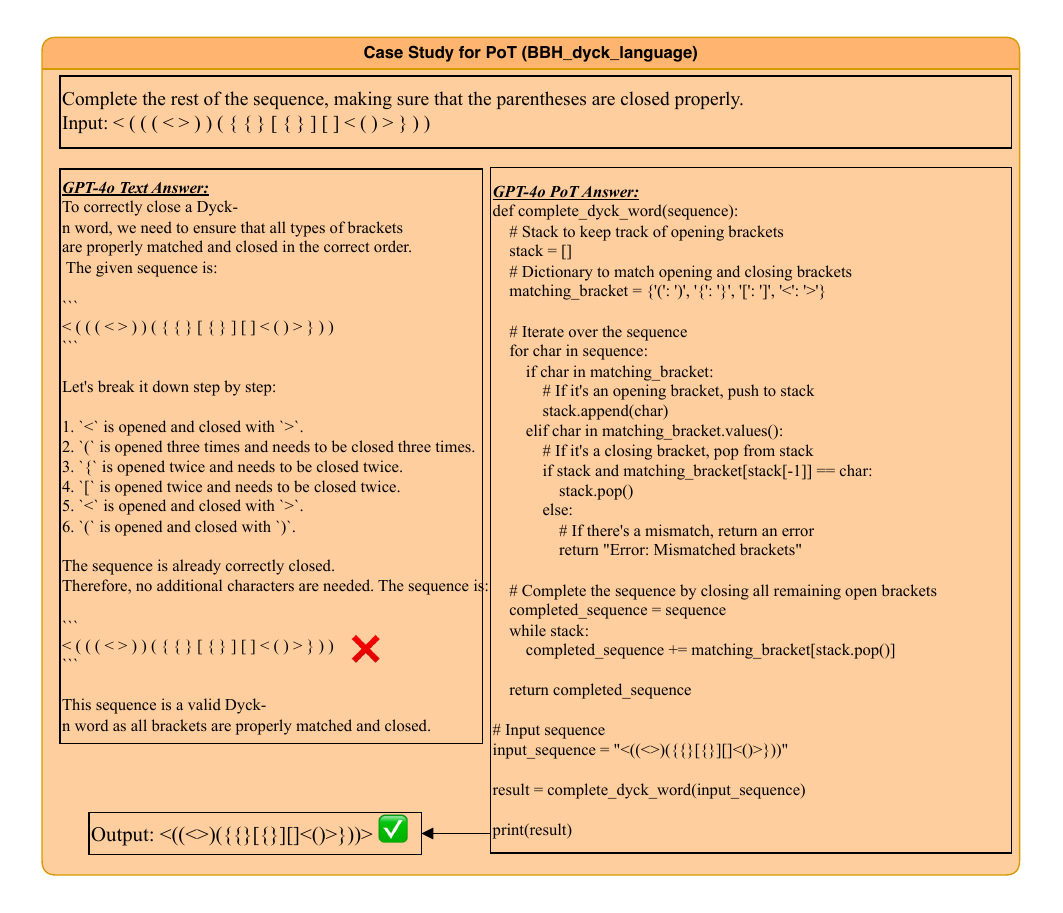}
    \caption{Case for PoT. Symbolic reasoning tasks are inherently well-suited to programming languages, and such problems may in fact originate from coding algorithm questions. Return to section \ref{sec: task_preference}}
    \label{fig:Case for PoT}
\end{figure*}
\begin{figure*}[tp]
    \centering
    \includegraphics[width=0.9\textwidth]{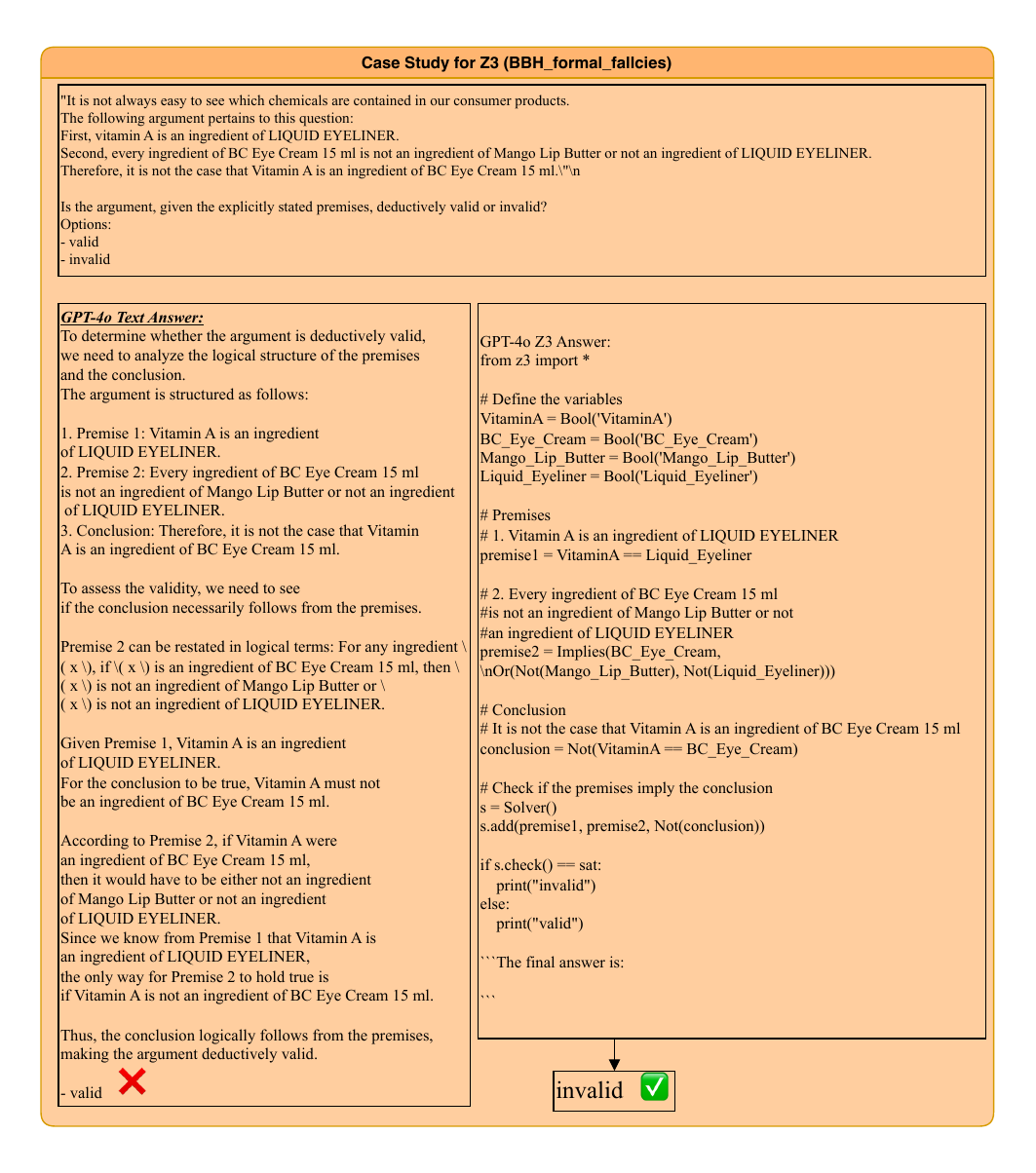}
    \caption{Case for Z3. Z3 (which, in this context, incorporates the first-order logic reasoning capabilities of Prover9) excels at solving formal first-order logic problems. Return to section \ref{sec: task_preference}}
    \label{fig:Case for Z3}
\end{figure*}
\begin{figure*}[tp]
    \centering
    \includegraphics[width=0.9\textwidth]{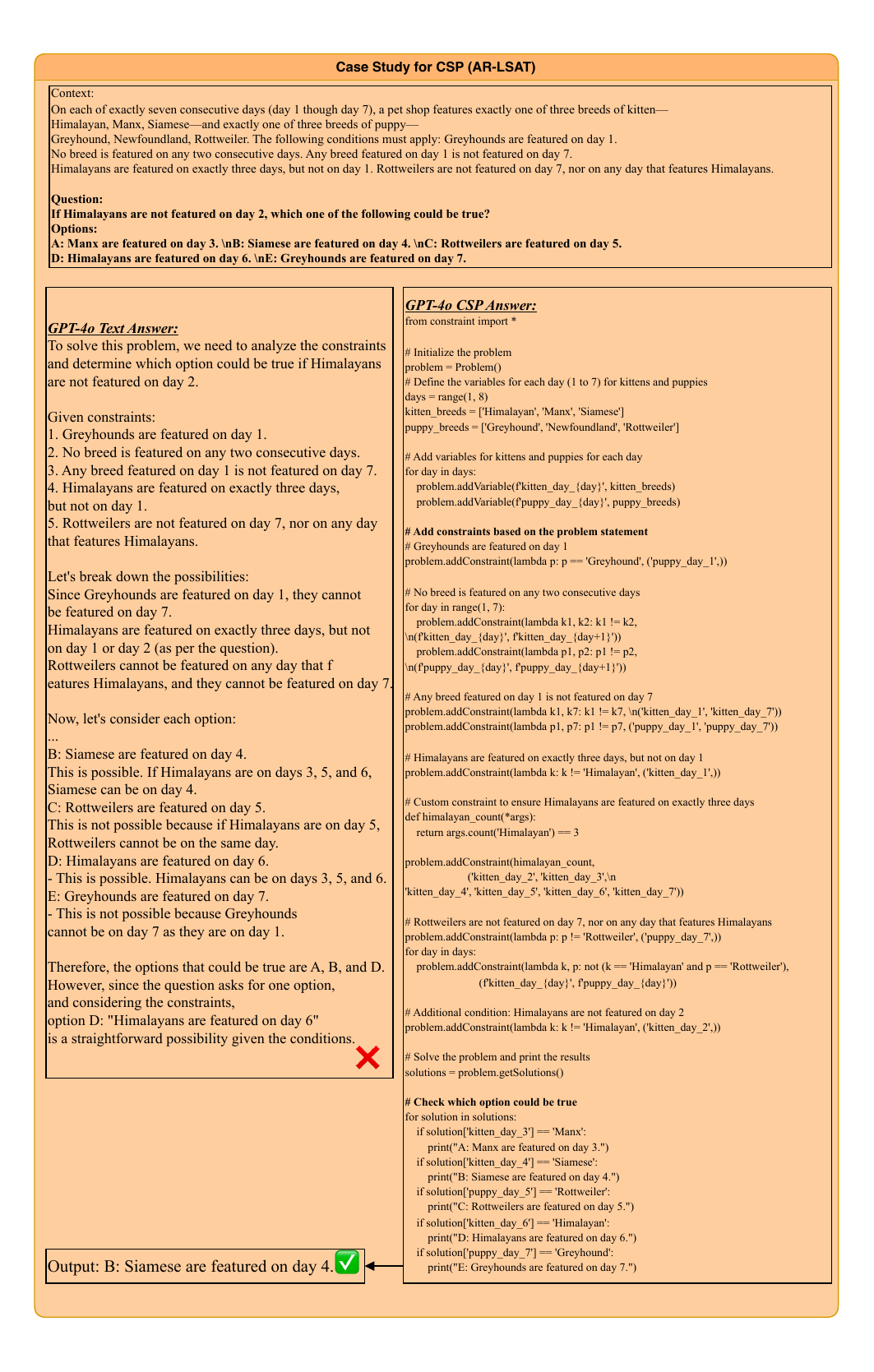}
    \caption{Case for CSP. The problems in ARLSAT involve numerous conditional constraints, which closely resemble the modeling approach used in Constraint Satisfaction Problems (CSPs). Return to section \ref{sec: task_preference}}
    \label{fig:Case for CSP}
\end{figure*}

\begin{figure*}[tp]
    \centering
    \includegraphics[width=0.9\textwidth]{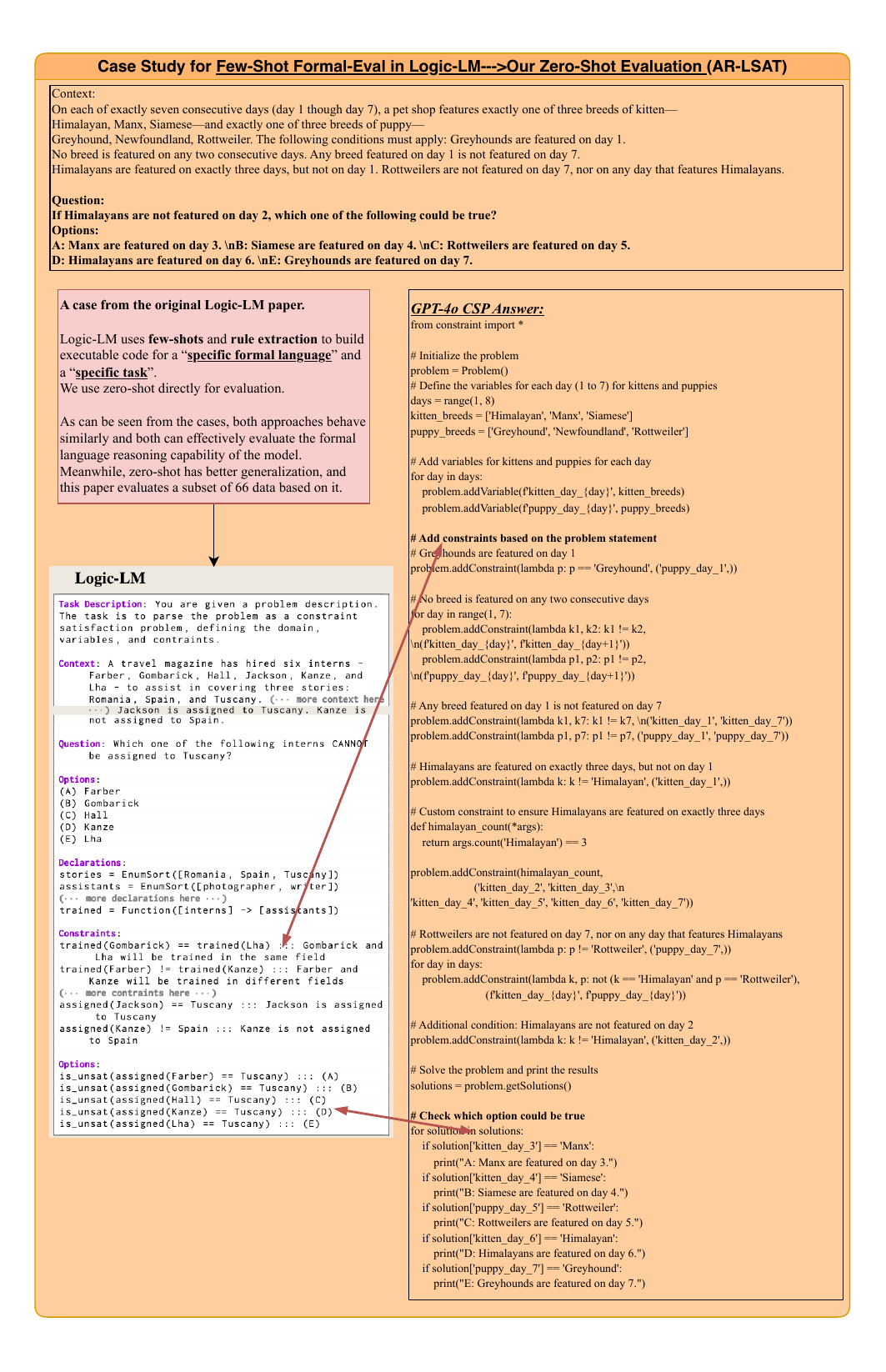}
    \caption{Case Study for Few-Shot Formal-Eval in Logic-LM--->Our Zero-Shot Evaluation (AR-LSAT). Logic-LM uses few-shots and rule extraction to build executable code for a “specific formal language” and a “specific task”. We use zero-shot directly for evaluation. As can be seen from the cases, both approaches behave similarly and both can effectively evaluate the formal language reasoning capability of the model. Meanwhile, zero-shot has better generalization, and this paper evaluates a subset of 66 data based on it.}
    \label{fig:case for compare}
\end{figure*}

\subsection{Case for PART I}
We give cases where Text (Fig~\ref{fig:Case for Text}), PoT(Fig~\ref{fig:Case for PoT}), Z3(Fig~\ref{fig:Case for Z3}) and CSP(Fig~\ref{fig:Case for CSP}) specialize in each case to show their strengths.

\subsection{From Logic-LM Few-Shot Eval to Zero-Shot}
Logic-LM uses few-shots setting and rule extraction to build “task-specific executable code” for “a particular formal language” and “a particular task”.
We use zero-shot directly for evaluation.
As shown in the case study in Figure~\ref{fig:case for compare}, both approaches behave similarly and can evaluate the model's formal language reasoning ability. Meanwhile, the zero-shot setting has better generalization, and this paper considers a subset of 66 datasets based on it.

\begin{table*}[htbp]
  \centering
  \resizebox{\linewidth}{!}{%
    \begin{tabular}{crrrrrrrrr}
    \toprule
    \multirow{2}[4]{*}{\textbf{Dataset}} & \multicolumn{1}{c}{\textbf{Text}} & \multicolumn{2}{c}{\textbf{PoT}} & \multicolumn{2}{c}{\textbf{Z3}} & \multicolumn{2}{c}{\textbf{CSP}} & \multicolumn{2}{c}{\textbf{AVG}} \\
\cmidrule{2-10}          & \multicolumn{1}{c}{\textbf{ACC}} & \multicolumn{1}{c}{\textbf{ACC}} & \multicolumn{1}{c}{\textbf{Exec\_Rate}} & \multicolumn{1}{c}{\textbf{ACC}} & \multicolumn{1}{c}{\textbf{Exec\_Rate}} & \multicolumn{1}{c}{\textbf{ACC}} & \multicolumn{1}{c}{\textbf{Exec\_Rate}} & \multicolumn{1}{c}{\textbf{ACC}} & \multicolumn{1}{c}{\textbf{Exec\_Rate}} \\
    \midrule
    \textbf{Average} & 75.0 & 68.6 & 85.1 & 61.9 & 79.4 & 65.1 & 82.2 & 67.6 & 82.2 \\
    \midrule
    \textbf{FOLIO} & 94.0 & 94.0 & 100.0 & 91.0 & 99.3 & 94.0 & 100.0 & 93.3 & 99.8 \\
    \textbf{ProntoQA} & 99.6 & 97.8 & 100.0 & 99.4 & 100.0 & 98.8 & 100.0 & 98.9 & 100.0 \\
    \textbf{logicbenchBQA} & 82.9 & 85.6 & 100.0 & 86.4 & 100.0 & 85.3 & 100.0 & 85.1 & 100.0 \\
    \textbf{BoardgameQA} & 78.5 & 79.3 & 99.9 & 75.1 & 100.0 & 69.6 & 100.0 & 75.6 & 100.0 \\
    \textbf{ARLSAT} & 92.2 & 91.3 & 100.0 & 83.0 & 97.0 & 89.1 & 100.0 & 88.9 & 99.0 \\
    \textbf{BBH\_boolean\_expressions} & 96.4 & 98.8 & 100.0 & 94.8 & 100.0 & 99.2 & 100.0 & 97.3 & 100.0 \\
    \textbf{bbeh\_boolean\_expressions} & 57.0 & 41.5 & 53.5 & 30.0 & 36.0 & 42.5 & 58.5 & 42.8 & 49.3 \\
    \textbf{BBH\_formal\_fallacies} & 100.0 & 99.2 & 100.0 & 99.6 & 99.6 & 98.8 & 100.0 & 99.4 & 99.9 \\
    \textbf{bbeh\_zebra\_puzzles} & 44.5 & 15.5 & 35.5 & 2.5 & 5.5 & 8.9 & 11.4 & 17.9 & 17.5 \\
    \textbf{BBH\_logical\_deduction\_five\_objects} & 100.0 & 100.0 & 100.0 & 99.6 & 100.0 & 98.0 & 100.0 & 99.4 & 100.0 \\
    \textbf{BBH\_logical\_deduction\_seven\_objects} & 99.2 & 99.6 & 100.0 & 100.0 & 100.0 & 100.0 & 100.0 & 99.7 & 100.0 \\
    \textbf{BBH\_logical\_deduction\_three\_objects} & 100.0 & 100.0 & 100.0 & 99.2 & 99.6 & 99.2 & 100.0 & 99.6 & 99.9 \\
    \textbf{bbeh\_boardgame\_qa} & 54.5 & 55.0 & 99.0 & 35.0 & 73.0 & 49.5 & 87.5 & 48.5 & 86.5 \\
    \textbf{BBH\_tracking\_shuffled\_objects\_five\_objects} & 100.0 & 100.0 & 100.0 & 98.8 & 99.6 & 98.0 & 100.0 & 99.2 & 99.9 \\
    \textbf{BBH\_tracking\_shuffled\_objects\_seven\_objects} & 100.0 & 100.0 & 100.0 & 96.8 & 100.0 & 99.2 & 100.0 & 99.0 & 100.0 \\
    \textbf{BBH\_tracking\_shuffled\_objects\_three\_objects} & 100.0 & 100.0 & 100.0 & 100.0 & 100.0 & 99.2 & 100.0 & 99.8 & 100.0 \\
    \textbf{bbeh\_shuffled\_objects} & 41.5 & 0.5 & 2.0 & 3.5 & 10.0 & 3.0 & 9.5 & 12.1 & 7.2 \\
    \textbf{BBH\_web\_of\_lies} & 92.8 & 98.8 & 100.0 & 98.0 & 99.6 & 99.2 & 100.0 & 97.2 & 99.9 \\
    \textbf{bbeh\_web\_of\_lies} & 58.0 & 37.5 & 43.5 & 12.0 & 17.0 & 21.5 & 24.5 & 32.3 & 28.3 \\
    \textbf{bAbI15} & 99.3 & 92.8 & 100.0 & 84.1 & 98.1 & 92.6 & 99.9 & 92.2 & 99.3 \\
    \textbf{NeuLRdeductive} & 99.9 & 97.3 & 100.0 & 80.9 & 98.2 & 95.8 & 100.0 & 93.5 & 99.4 \\
    \textbf{clutrr} & 78.8 & 73.3 & 100.0 & 60.1 & 94.2 & 71.0 & 98.7 & 70.8 & 97.6 \\
    \textbf{bAbI16} & 85.5 & 91.8 & 100.0 & 92.1 & 100.0 & 89.7 & 100.0 & 89.8 & 100.0 \\
    \textbf{NeuLRinductive} & 76.3 & 73.3 & 99.9 & 90.1 & 99.6 & 80.7 & 99.8 & 80.1 & 99.8 \\
    \textbf{anli} & 86.8 & 86.9 & 100.0 & 81.3 & 99.9 & 85.9 & 99.9 & 85.2 & 99.9 \\
    \textbf{AbductionRules} & 68.8 & 71.5 & 100.0 & 45.5 & 98.8 & 62.8 & 94.0 & 62.2 & 97.6 \\
    \textbf{BBH\_causal\_judgement} & 64.2 & 64.7 & 100.0 & 59.4 & 100.0 & 64.2 & 100.0 & 63.1 & 100.0 \\
    \textbf{bbeh\_causal\_understanding} & 62.0 & 53.5 & 99.5 & 46.5 & 90.5 & 49.0 & 94.5 & 52.8 & 94.8 \\
    \textbf{NeuLRabductive} & 26.0 & 26.9 & 99.9 & 9.9 & 95.7 & 15.1 & 94.1 & 19.5 & 96.6 \\
    \textbf{logicqa} & 86.5 & 82.9 & 100.0 & 77.9 & 99.6 & 80.5 & 99.9 & 82.0 & 99.8 \\
    \textbf{BBH\_date\_understanding} & 96.8 & 94.8 & 100.0 & 88.0 & 98.8 & 89.6 & 100.0 & 92.3 & 99.6 \\
    \textbf{bbeh\_time\_arithmetic} & 86.5 & 79.5 & 87.5 & 42.5 & 50.5 & 61.5 & 72.5 & 67.5 & 70.2 \\
    \textbf{BBH\_temporal\_sequences} & 100.0 & 99.6 & 100.0 & 91.6 & 99.2 & 97.2 & 99.2 & 97.1 & 99.5 \\
    \textbf{bbeh\_temporal\_sequence} & 52.5 & 0.0 & 0.5 & 0.0 & 0.0 & 1.0 & 1.5 & 13.4 & 0.7 \\
    \textbf{BBH\_disambiguation\_qa} & 48.0 & 54.0 & 100.0 & 38.8 & 100.0 & 46.4 & 100.0 & 46.8 & 100.0 \\
    \textbf{bbeh\_disambiguation\_qa} & 58.3 & 50.8 & 97.5 & 40.8 & 82.5 & 51.7 & 86.7 & 50.4 & 88.9 \\
    \textbf{BBH\_hyperbaton} & 100.0 & 100.0 & 100.0 & 99.6 & 100.0 & 96.4 & 100.0 & 99.0 & 100.0 \\
    \textbf{bbeh\_hyperbaton} & 38.0 & 26.5 & 56.5 & 19.0 & 35.0 & 18.0 & 52.0 & 25.4 & 47.8 \\
    \textbf{BBH\_ruin\_names} & 80.0 & 84.4 & 100.0 & 78.4 & 96.4 & 82.8 & 100.0 & 81.4 & 98.8 \\
    \textbf{bbeh\_nycc} & 15.0 & 8.5 & 72.0 & 11.5 & 82.5 & 10.5 & 75.0 & 11.4 & 76.5 \\
    \textbf{BBH\_salient\_translation\_error\_detection} & 76.8 & 74.8 & 100.0 & 76.0 & 99.6 & 75.6 & 99.6 & 75.8 & 99.7 \\
    \textbf{bbeh\_linguini} & 46.0 & 22.0 & 79.5 & 24.5 & 65.5 & 17.5 & 51.0 & 27.5 & 65.3 \\
    \textbf{BBH\_snarks} & 93.8 & 91.6 & 98.9 & 90.5 & 100.0 & 92.1 & 100.0 & 92.0 & 99.6 \\
    \textbf{bbeh\_sarc\_triples} & 32.0 & 39.0 & 100.0 & 25.5 & 71.5 & 19.0 & 83.0 & 28.9 & 84.8 \\
    \textbf{BBH\_dyck\_languages} & 91.6 & 73.6 & 83.2 & 72.4 & 92.0 & 84.0 & 98.8 & 80.4 & 91.3 \\
    \textbf{bbeh\_dyck\_languages} & 49.0 & 30.0 & 79.5 & 22.0 & 69.0 & 28.0 & 67.5 & 32.3 & 72.0 \\
    \textbf{BBH\_word\_sorting} & 98.8 & 100.0 & 100.0 & 20.4 & 21.2 & 79.2 & 88.0 & 74.6 & 69.7 \\
    \textbf{bbeh\_word\_sorting} & 77.0 & 77.5 & 92.5 & 63.5 & 76.0 & 40.0 & 62.0 & 64.5 & 76.8 \\
    \textbf{BBH\_geometric\_shapes} & 80.0 & 80.8 & 100.0 & 82.0 & 99.2 & 78.4 & 100.0 & 80.3 & 99.7 \\
    \textbf{bbeh\_geometric\_shapes} & 40.5 & 9.0 & 19.5 & 12.5 & 25.5 & 14.5 & 32.0 & 19.1 & 25.7 \\
    \textbf{BBH\_navigate} & 97.6 & 98.4 & 100.0 & 93.6 & 99.6 & 95.2 & 99.6 & 96.2 & 99.7 \\
    \textbf{bbeh\_spatial\_reasoning} & 43.0 & 40.5 & 53.0 & 32.5 & 48.0 & 30.5 & 47.5 & 36.6 & 49.5 \\
    \textbf{BBH\_penguins\_in\_a\_table} & 99.3 & 99.3 & 100.0 & 95.2 & 95.2 & 99.3 & 100.0 & 98.3 & 98.4 \\
    \textbf{bbeh\_buggy\_tables} & 25.0 & 16.5 & 28.5 & 8.0 & 16.0 & 3.5 & 15.0 & 13.3 & 19.8 \\
    \textbf{BBH\_movie\_recommendation} & 70.0 & 70.0 & 100.0 & 63.6 & 97.6 & 65.2 & 100.0 & 67.2 & 99.2 \\
    \textbf{bbeh\_movie\_recommendation} & 59.5 & 40.0 & 79.0 & 28.5 & 56.5 & 22.5 & 47.5 & 37.6 & 61.0 \\
    \textbf{BBH\_sports\_understanding} & 81.6 & 80.4 & 100.0 & 77.2 & 99.6 & 80.0 & 100.0 & 79.8 & 99.9 \\
    \textbf{bbeh\_sportqa} & 53.0 & 15.0 & 23.0 & 17.0 & 34.0 & 17.5 & 34.5 & 25.6 & 30.5 \\
    \textbf{gsm8k} & 96.4 & 96.7 & 99.9 & 93.2 & 98.8 & 94.5 & 99.5 & 95.2 & 99.4 \\
    \textbf{MATH} & 96.5 & 93.1 & 99.4 & 70.7 & 88.3 & 87.1 & 99.0 & 86.9 & 95.6 \\
    \textbf{BBH\_multistep\_arithmetic\_two} & 100.0 & 100.0 & 100.0 & 100.0 & 100.0 & 99.6 & 99.6 & 99.9 & 99.9 \\
    \textbf{bbeh\_multistep\_arithmetic} & 53.0 & 38.0 & 40.5 & 20.5 & 29.5 & 22.7 & 32.8 & 33.6 & 34.3 \\
    \textbf{BBH\_object\_counting} & 100.0 & 100.0 & 100.0 & 98.8 & 99.6 & 98.4 & 99.6 & 99.3 & 99.7 \\
    \textbf{bbeh\_object\_counting} & 58.0 & 31.0 & 85.5 & 40.5 & 71.0 & 22.5 & 61.0 & 38.0 & 72.5 \\
    \textbf{BBH\_reasoning\_about\_colored\_objects} & 100.0 & 98.8 & 100.0 & 95.6 & 99.2 & 94.4 & 99.2 & 97.2 & 99.5 \\
    \textbf{bbeh\_object\_properties} & 31.5 & 5.5 & 7.0 & 17.5 & 29.5 & 39.0 & 51.5 & 23.4 & 29.3 \\
    \bottomrule
    \end{tabular}}
    \caption{QwQ-32B Full Result.}
  \label{tab:QwQ-32B}%
\end{table*}%

\begin{table*}[htbp]
  \centering
  \resizebox{!}{12cm}{%
    \begin{tabular}{crrrrrrrrr}
    \toprule
    \multirow{2}[4]{*}{\textbf{Dataset}} & \multicolumn{1}{c}{\textbf{Text}} & \multicolumn{2}{c}{\textbf{PoT}} & \multicolumn{2}{c}{\textbf{Z3}} & \multicolumn{2}{c}{\textbf{CSP}} & \multicolumn{2}{c}{\textbf{AVG}} \\
\cmidrule{2-10}          & \multicolumn{1}{c}{\textbf{ACC}} & \multicolumn{1}{c}{\textbf{ACC}} & \multicolumn{1}{c}{\textbf{Exec\_Rate}} & \multicolumn{1}{c}{\textbf{ACC}} & \multicolumn{1}{c}{\textbf{Exec\_Rate}} & \multicolumn{1}{c}{\textbf{ACC}} & \multicolumn{1}{c}{\textbf{Exec\_Rate}} & \multicolumn{1}{c}{\textbf{ACC}} & \multicolumn{1}{c}{\textbf{Exec\_Rate}} \\
    \midrule
    \textbf{Average} & 66.7 & 63.5 & 91.5 & 54.5 & 87.4 & 52.8 & 84.0 & 59.4 & 87.6 \\
    \midrule
    \textbf{FOLIO} & 92.5 & 88.1 & 100.0 & 73.9 & 88.8 & 67.2 & 98.5 & 80.4 & 95.8 \\
    \textbf{ProntoQA} & 100.0 & 95.8 & 100.0 & 80.2 & 99.8 & 93.2 & 99.4 & 92.3 & 99.7 \\
    \textbf{logicbenchBQA} & 72.3 & 70.5 & 99.8 & 76.3 & 100.0 & 63.3 & 99.8 & 70.6 & 99.9 \\
    \textbf{BoardgameQA} & 59.7 & 66.0 & 100.0 & 63.5 & 97.9 & 60.2 & 98.1 & 62.4 & 98.7 \\
    \textbf{ARLSAT} & 40.9 & 50.4 & 90.4 & 59.1 & 83.0 & 67.8 & 85.2 & 54.6 & 86.2 \\
    \textbf{BBH\_boolean\_expressions} & 99.6 & 100.0 & 100.0 & 89.2 & 100.0 & 76.4 & 96.4 & 91.3 & 98.8 \\
    \textbf{bbeh\_boolean\_expressions} & 59.5 & 51.5 & 55.0 & 1.5 & 2.0 & 56.5 & 60.0 & 42.3 & 39.0 \\
    \textbf{BBH\_formal\_fallacies} & 88.4 & 76.4 & 100.0 & 90.4 & 100.0 & 62.0 & 99.6 & 79.3 & 99.9 \\
    \textbf{bbeh\_zebra\_puzzles} & 38.0 & 6.5 & 19.0 & 8.5 & 49.0 & 3.0 & 4.0 & 14.0 & 24.0 \\
    \textbf{BBH\_logical\_deduction\_five\_objects} & 93.2 & 94.0 & 99.6 & 87.6 & 99.6 & 96.4 & 100.0 & 92.8 & 99.7 \\
    \textbf{BBH\_logical\_deduction\_seven\_objects} & 88.8 & 88.0 & 100.0 & 84.8 & 100.0 & 96.8 & 100.0 & 89.6 & 100.0 \\
    \textbf{BBH\_logical\_deduction\_three\_objects} & 99.2 & 95.2 & 100.0 & 92.8 & 99.2 & 99.6 & 100.0 & 96.7 & 99.7 \\
    \textbf{bbeh\_boardgame\_qa} & 37.0 & 35.5 & 90.5 & 37.5 & 90.0 & 24.5 & 65.0 & 33.6 & 81.8 \\
    \textbf{BBH\_tracking\_shuffled\_objects\_five\_objects} & 98.4 & 100.0 & 100.0 & 82.0 & 100.0 & 36.8 & 81.2 & 79.3 & 93.7 \\
    \textbf{BBH\_tracking\_shuffled\_objects\_seven\_objects} & 100.0 & 99.6 & 100.0 & 82.4 & 100.0 & 41.6 & 75.2 & 80.9 & 91.7 \\
    \textbf{BBH\_tracking\_shuffled\_objects\_three\_objects} & 100.0 & 100.0 & 100.0 & 55.6 & 100.0 & 53.6 & 74.4 & 77.3 & 91.5 \\
    \textbf{bbeh\_shuffled\_objects} & 29.5 & 59.0 & 83.5 & 36.0 & 77.5 & 23.5 & 49.0 & 37.0 & 70.0 \\
    \textbf{BBH\_web\_of\_lies} & 96.4 & 91.2 & 100.0 & 96.4 & 100.0 & 96.4 & 100.0 & 95.1 & 100.0 \\
    \textbf{bbeh\_web\_of\_lies} & 33.5 & 11.0 & 51.5 & 11.0 & 20.5 & 11.5 & 14.5 & 16.8 & 28.8 \\
    \textbf{bAbI15} & 99.6 & 98.7 & 100.0 & 76.2 & 97.9 & 95.9 & 99.8 & 92.6 & 99.2 \\
    \textbf{NeuLRdeductive} & 99.8 & 97.0 & 100.0 & 55.2 & 93.9 & 87.2 & 97.8 & 84.8 & 97.2 \\
    \textbf{clutrr} & 52.7 & 44.2 & 100.0 & 44.6 & 95.7 & 35.6 & 84.4 & 44.3 & 93.4 \\
    \textbf{bAbI16} & 51.8 & 93.4 & 100.0 & 64.4 & 98.8 & 44.1 & 100.0 & 63.4 & 99.6 \\
    \textbf{NeuLRinductive} & 60.3 & 41.2 & 100.0 & 21.1 & 99.3 & 7.9 & 97.2 & 32.6 & 98.8 \\
    \textbf{anli} & 88.8 & 87.6 & 100.0 & 73.4 & 99.9 & 81.6 & 100.0 & 82.9 & 100.0 \\
    \textbf{AbductionRules} & 88.5 & 86.6 & 100.0 & 84.3 & 100.0 & 41.2 & 63.8 & 75.2 & 87.9 \\
    \textbf{BBH\_causal\_judgement} & 69.0 & 73.8 & 100.0 & 61.0 & 100.0 & 64.7 & 100.0 & 67.1 & 100.0 \\
    \textbf{bbeh\_causal\_understanding} & 52.0 & 52.5 & 100.0 & 50.0 & 99.0 & 44.5 & 96.5 & 49.8 & 98.5 \\
    \textbf{NeuLRabductive} & 29.0 & 15.0 & 98.4 & 19.8 & 92.6 & 5.2 & 88.4 & 17.3 & 93.1 \\
    \textbf{logicqa} & 76.0 & 73.2 & 99.6 & 61.7 & 97.8 & 72.7 & 98.6 & 70.9 & 98.7 \\
    \textbf{BBH\_date\_understanding} & 94.0 & 82.0 & 100.0 & 70.4 & 98.8 & 84.4 & 100.0 & 82.7 & 99.6 \\
    \textbf{bbeh\_time\_arithmetic} & 63.5 & 43.5 & 74.0 & 36.0 & 66.0 & 35.0 & 73.5 & 44.5 & 71.2 \\
    \textbf{BBH\_temporal\_sequences} & 99.6 & 89.2 & 99.6 & 64.4 & 98.0 & 98.0 & 99.6 & 87.8 & 99.1 \\
    \textbf{bbeh\_temporal\_sequence} & 5.5 & 2.0 & 87.5 & 1.0 & 52.0 & 2.5 & 81.5 & 2.8 & 73.7 \\
    \textbf{BBH\_disambiguation\_qa} & 53.6 & 50.0 & 100.0 & 38.0 & 100.0 & 36.8 & 100.0 & 44.6 & 100.0 \\
    \textbf{bbeh\_disambiguation\_qa} & 63.3 & 54.2 & 98.3 & 44.2 & 96.7 & 70.8 & 93.3 & 58.1 & 96.1 \\
    \textbf{BBH\_hyperbaton} & 92.8 & 88.4 & 100.0 & 94.8 & 99.6 & 92.0 & 99.2 & 92.0 & 99.6 \\
    \textbf{bbeh\_hyperbaton} & 30.5 & 13.0 & 87.0 & 28.0 & 90.0 & 17.0 & 42.5 & 22.1 & 73.2 \\
    \textbf{BBH\_ruin\_names} & 86.4 & 86.0 & 100.0 & 80.8 & 98.4 & 84.4 & 99.2 & 84.4 & 99.2 \\
    \textbf{bbeh\_nycc} & 21.5 & 11.5 & 77.5 & 3.5 & 39.0 & 7.0 & 79.0 & 10.9 & 65.2 \\
    \textbf{BBH\_salient\_translation\_error\_detection} & 73.2 & 80.0 & 100.0 & 84.4 & 100.0 & 76.8 & 100.0 & 78.6 & 100.0 \\
    \textbf{bbeh\_linguini} & 35.0 & 25.5 & 97.0 & 23.0 & 92.0 & 20.0 & 73.0 & 25.9 & 87.3 \\
    \textbf{BBH\_snarks} & 89.9 & 86.5 & 100.0 & 75.3 & 100.0 & 74.7 & 100.0 & 81.6 & 100.0 \\
    \textbf{bbeh\_sarc\_triples} & 30.0 & 37.5 & 97.5 & 17.5 & 45.5 & 35.5 & 85.0 & 30.1 & 76.0 \\
    \textbf{BBH\_dyck\_languages} & 90.4 & 94.4 & 100.0 & 75.2 & 98.8 & 35.2 & 76.0 & 73.8 & 91.6 \\
    \textbf{bbeh\_dyck\_languages} & 17.5 & 6.5 & 86.5 & 7.5 & 96.0 & 10.0 & 88.0 & 10.4 & 90.2 \\
    \textbf{BBH\_word\_sorting} & 97.6 & 99.2 & 100.0 & 41.2 & 94.0 & 71.6 & 84.0 & 77.4 & 92.7 \\
    \textbf{bbeh\_word\_sorting} & 46.5 & 53.0 & 89.5 & 43.5 & 83.5 & 30.0 & 61.5 & 43.3 & 78.2 \\
    \textbf{BBH\_geometric\_shapes} & 71.2 & 77.2 & 99.6 & 83.2 & 100.0 & 77.6 & 100.0 & 77.3 & 99.9 \\
    \textbf{bbeh\_geometric\_shapes} & 38.5 & 16.0 & 65.0 & 35.0 & 80.5 & 33.5 & 69.5 & 30.8 & 71.7 \\
    \textbf{BBH\_navigate} & 99.2 & 98.0 & 100.0 & 78.4 & 100.0 & 74.4 & 98.0 & 87.5 & 99.3 \\
    \textbf{bbeh\_spatial\_reasoning} & 11.5 & 19.0 & 81.0 & 13.5 & 80.5 & 14.5 & 72.0 & 14.6 & 77.8 \\
    \textbf{BBH\_penguins\_in\_a\_table} & 98.6 & 100.0 & 100.0 & 96.6 & 99.3 & 82.2 & 97.3 & 94.4 & 98.9 \\
    \textbf{bbeh\_buggy\_tables} & 21.0 & 19.0 & 49.5 & 16.5 & 54.0 & 19.5 & 40.0 & 19.0 & 47.8 \\
    \textbf{BBH\_movie\_recommendation} & 77.2 & 63.6 & 100.0 & 67.2 & 99.2 & 76.8 & 99.6 & 71.2 & 99.6 \\
    \textbf{bbeh\_movie\_recommendation} & 60.5 & 26.0 & 99.5 & 34.0 & 85.0 & 27.5 & 85.0 & 37.0 & 89.8 \\
    \textbf{BBH\_sports\_understanding} & 86.8 & 87.6 & 100.0 & 54.4 & 100.0 & 60.8 & 100.0 & 72.4 & 100.0 \\
    \textbf{bbeh\_sportqa} & 29.5 & 58.5 & 100.0 & 23.5 & 89.5 & 26.0 & 85.5 & 34.4 & 91.7 \\
    \textbf{gsm8k} & 96.7 & 94.6 & 98.8 & 89.3 & 98.3 & 87.6 & 99.3 & 92.1 & 98.8 \\
    \textbf{MATH} & 81.3 & 69.3 & 88.7 & 57.9 & 91.2 & 57.7 & 83.5 & 66.6 & 87.8 \\
    \textbf{BBH\_multistep\_arithmetic\_two} & 98.8 & 100.0 & 100.0 & 81.2 & 99.2 & 63.6 & 100.0 & 85.9 & 99.7 \\
    \textbf{bbeh\_multistep\_arithmetic} & 27.0 & 1.5 & 23.0 & 1.0 & 13.5 & 1.0 & 16.0 & 7.6 & 17.5 \\
    \textbf{BBH\_object\_counting} & 95.6 & 99.6 & 100.0 & 98.8 & 100.0 & 88.0 & 100.0 & 95.5 & 100.0 \\
    \textbf{bbeh\_object\_counting} & 16.0 & 14.0 & 98.5 & 11.0 & 97.5 & 13.0 & 97.0 & 13.5 & 97.7 \\
    \textbf{BBH\_reasoning\_about\_colored\_objects} & 97.6 & 97.6 & 100.0 & 94.4 & 98.8 & 89.2 & 99.6 & 94.7 & 99.5 \\
    \textbf{bbeh\_object\_properties} & 10.5 & 5.5 & 50.5 & 9.5 & 52.5 & 1.0 & 38.5 & 6.6 & 47.2 \\
    \bottomrule
    \end{tabular}}
    \caption{GPT-4o Full Result.}
  \label{tab:GPT-4o}%
\end{table*}%

\begin{table*}[htbp]
  \centering
  \resizebox{!}{12cm}{%
    \begin{tabular}{crrrrrrrrr}
    \toprule
    \multirow{2}[4]{*}{\textbf{Dataset}} & \multicolumn{1}{c}{\textbf{Text}} & \multicolumn{2}{c}{\textbf{PoT}} & \multicolumn{2}{c}{\textbf{Z3}} & \multicolumn{2}{c}{\textbf{CSP}} & \multicolumn{2}{c}{\textbf{AVG}} \\
\cmidrule{2-10}          & \multicolumn{1}{c}{\textbf{ACC}} & \multicolumn{1}{c}{\textbf{ACC}} & \multicolumn{1}{c}{\textbf{Exec\_Rate}} & \multicolumn{1}{c}{\textbf{ACC}} & \multicolumn{1}{c}{\textbf{Exec\_Rate}} & \multicolumn{1}{c}{\textbf{ACC}} & \multicolumn{1}{c}{\textbf{Exec\_Rate}} & \multicolumn{1}{c}{\textbf{ACC}} & \multicolumn{1}{c}{\textbf{Exec\_Rate}} \\
    \midrule
    \textbf{Average} & 52.3 & 36.9 & 78.6 & 33.0 & 70.0 & 24.8 & 52.1 & 36.7 & 66.9 \\
    \midrule
    \textbf{FOLIO} & 88.8 & 85.1 & 100.0 & 59.7 & 87.3 & 59.0 & 84.3 & 73.1 & 90.5 \\
    \textbf{ProntoQA} & 99.4 & 83.4 & 98.6 & 57.2 & 87.6 & 38.4 & 54.8 & 69.6 & 80.3 \\
    \textbf{logicbenchBQA} & 71.6 & 52.1 & 100.0 & 39.4 & 98.5 & 37.8 & 79.8 & 50.2 & 92.8 \\
    \textbf{BoardgameQA} & 54.3 & 52.8 & 99.1 & 38.2 & 89.5 & 29.2 & 73.8 & 43.6 & 87.5 \\
    \textbf{ARLSAT} & 25.2 & 36.5 & 93.9 & 22.2 & 51.3 & 8.7 & 22.6 & 23.2 & 55.9 \\
    \textbf{BBH\_boolean\_expressions} & 97.6 & 99.6 & 100.0 & 45.2 & 71.6 & 50.8 & 95.2 & 73.3 & 88.9 \\
    \textbf{bbeh\_boolean\_expressions} & 70.5 & 1.5 & 1.5 & 0.5 & 0.5 & 10.0 & 10.5 & 20.6 & 4.2 \\
    \textbf{BBH\_formal\_fallacies} & 69.6 & 50.8 & 100.0 & 52.8 & 91.6 & 50.0 & 88.4 & 55.8 & 93.3 \\
    \textbf{bbeh\_zebra\_puzzles} & 34.5 & 0.5 & 3.5 & 5.0 & 20.5 & 0.0 & 0.0 & 10.0 & 8.0 \\
    \textbf{BBH\_logical\_deduction\_five\_objects} & 66.8 & 56.0 & 100.0 & 46.8 & 69.2 & 64.8 & 95.2 & 58.6 & 88.1 \\
    \textbf{BBH\_logical\_deduction\_seven\_objects} & 66.0 & 55.2 & 100.0 & 47.2 & 67.6 & 70.8 & 94.8 & 59.8 & 87.5 \\
    \textbf{BBH\_logical\_deduction\_three\_objects} & 89.6 & 74.4 & 100.0 & 48.8 & 73.2 & 72.4 & 92.4 & 71.3 & 88.5 \\
    \textbf{bbeh\_boardgame\_qa} & 33.0 & 18.5 & 46.5 & 7.0 & 20.5 & 0.5 & 5.0 & 14.8 & 24.0 \\
    \textbf{BBH\_tracking\_shuffled\_objects\_five\_objects} & 84.8 & 3.6 & 100.0 & 34.8 & 82.0 & 16.8 & 63.6 & 35.0 & 81.9 \\
    \textbf{BBH\_tracking\_shuffled\_objects\_seven\_objects} & 85.2 & 5.2 & 100.0 & 43.2 & 82.0 & 15.6 & 68.4 & 37.3 & 83.5 \\
    \textbf{BBH\_tracking\_shuffled\_objects\_three\_objects} & 89.2 & 0.4 & 100.0 & 35.6 & 76.4 & 22.0 & 58.4 & 36.8 & 78.3 \\
    \textbf{bbeh\_shuffled\_objects} & 59.5 & 4.0 & 26.5 & 2.0 & 12.5 & 1.5 & 4.0 & 16.8 & 14.3 \\
    \textbf{BBH\_web\_of\_lies} & 81.2 & 59.2 & 100.0 & 78.4 & 94.4 & 66.8 & 74.8 & 71.4 & 89.7 \\
    \textbf{bbeh\_web\_of\_lies} & 9.0 & 4.0 & 13.0 & 1.0 & 5.0 & 0.5 & 3.5 & 3.6 & 7.2 \\
    \textbf{bAbI15} & 23.7 & 54.3 & 99.9 & 29.1 & 90.7 & 16.0 & 64.2 & 30.8 & 84.9 \\
    \textbf{NeuLRdeductive} & 91.9 & 60.4 & 96.5 & 20.4 & 77.0 & 7.7 & 41.8 & 45.1 & 71.8 \\
    \textbf{clutrr} & 17.7 & 26.4 & 99.9 & 14.2 & 82.8 & 12.1 & 60.0 & 17.6 & 80.9 \\
    \textbf{bAbI16} & 23.7 & 55.8 & 99.9 & 31.3 & 91.6 & 14.8 & 63.5 & 31.4 & 85.0 \\
    \textbf{NeuLRinductive} & 7.4 & 8.8 & 96.5 & 16.1 & 91.1 & 14.4 & 53.3 & 11.7 & 80.3 \\
    \textbf{anli} & 77.7 & 78.8 & 99.8 & 59.8 & 95.1 & 55.6 & 83.7 & 68.0 & 92.9 \\
    \textbf{AbductionRules} & 88.3 & 50.6 & 81.4 & 34.8 & 41.8 & 23.9 & 37.0 & 49.4 & 53.4 \\
    \textbf{BBH\_causal\_judgement} & 51.9 & 54.0 & 100.0 & 37.4 & 92.5 & 40.6 & 85.6 & 46.0 & 92.7 \\
    \textbf{bbeh\_causal\_understanding} & 45.0 & 39.0 & 98.0 & 26.5 & 82.0 & 26.5 & 69.0 & 34.3 & 83.0 \\
    \textbf{NeuLRabductive} & 20.8 & 12.9 & 83.5 & 22.0 & 52.5 & 8.2 & 21.6 & 16.0 & 52.5 \\
    \textbf{logicqa} & 68.2 & 64.8 & 98.2 & 54.9 & 95.9 & 40.8 & 82.5 & 57.2 & 92.2 \\
    \textbf{BBH\_date\_understanding} & 84.8 & 32.0 & 100.0 & 33.6 & 73.6 & 38.4 & 84.8 & 47.2 & 86.1 \\
    \textbf{bbeh\_time\_arithmetic} & 30.5 & 3.0 & 32.0 & 9.0 & 52.5 & 10.5 & 46.0 & 13.3 & 43.5 \\
    \textbf{BBH\_temporal\_sequences} & 83.6 & 67.2 & 100.0 & 48.4 & 85.6 & 54.0 & 72.8 & 63.3 & 86.1 \\
    \textbf{bbeh\_temporal\_sequence} & 5.0 & 0.0 & 34.5 & 0.0 & 41.5 & 0.0 & 5.0 & 1.3 & 27.0 \\
    \textbf{BBH\_disambiguation\_qa} & 41.2 & 59.2 & 100.0 & 37.2 & 93.6 & 37.2 & 88.0 & 43.7 & 93.9 \\
    \textbf{bbeh\_disambiguation\_qa} & 45.8 & 29.2 & 90.0 & 35.8 & 81.7 & 15.8 & 44.2 & 31.7 & 71.9 \\
    \textbf{BBH\_hyperbaton} & 68.0 & 70.0 & 100.0 & 53.6 & 94.4 & 32.4 & 57.6 & 56.0 & 84.0 \\
    \textbf{bbeh\_hyperbaton} & 0.5 & 1.5 & 22.0 & 1.0 & 27.0 & 0.0 & 0.5 & 0.8 & 16.5 \\
    \textbf{BBH\_ruin\_names} & 53.2 & 36.0 & 100.0 & 28.4 & 82.8 & 10.4 & 39.2 & 32.0 & 74.0 \\
    \textbf{bbeh\_nycc} & 10.5 & 8.5 & 94.0 & 7.0 & 59.0 & 4.5 & 41.5 & 7.6 & 64.8 \\
    \textbf{BBH\_salient\_translation\_error\_detection} & 47.2 & 10.0 & 100.0 & 30.4 & 58.0 & 21.6 & 60.0 & 27.3 & 72.7 \\
    \textbf{bbeh\_linguini} & 18.0 & 16.0 & 81.5 & 6.0 & 73.5 & 4.5 & 52.0 & 11.1 & 69.0 \\
    \textbf{BBH\_snarks} & 77.5 & 31.5 & 100.0 & 49.4 & 97.2 & 41.0 & 84.8 & 49.9 & 94.0 \\
    \textbf{bbeh\_sarc\_triples} & 16.0 & 12.5 & 91.5 & 12.0 & 74.5 & 5.5 & 22.5 & 11.5 & 62.8 \\
    \textbf{BBH\_dyck\_languages} & 83.2 & 39.2 & 100.0 & 24.8 & 76.8 & 14.8 & 40.0 & 40.5 & 72.3 \\
    \textbf{bbeh\_dyck\_languages} & 4.5 & 1.5 & 26.0 & 8.0 & 57.5 & 0.5 & 26.0 & 3.6 & 36.5 \\
    \textbf{BBH\_word\_sorting} & 32.4 & 97.2 & 100.0 & 61.2 & 86.0 & 6.0 & 32.4 & 49.2 & 72.8 \\
    \textbf{bbeh\_word\_sorting} & 21.0 & 10.0 & 58.5 & 23.0 & 76.0 & 10.5 & 27.5 & 16.1 & 54.0 \\
    \textbf{BBH\_geometric\_shapes} & 62.0 & 80.4 & 100.0 & 48.4 & 86.0 & 55.2 & 89.2 & 61.5 & 91.7 \\
    \textbf{bbeh\_geometric\_shapes} & 28.5 & 19.5 & 53.5 & 30.0 & 98.5 & 30.0 & 65.0 & 27.0 & 72.3 \\
    \textbf{BBH\_navigate} & 84.0 & 50.4 & 100.0 & 43.6 & 86.0 & 29.6 & 50.8 & 51.9 & 78.9 \\
    \textbf{bbeh\_spatial\_reasoning} & 6.0 & 1.5 & 22.0 & 4.0 & 27.5 & 4.0 & 19.0 & 3.9 & 22.8 \\
    \textbf{BBH\_penguins\_in\_a\_table} & 91.1 & 56.2 & 100.0 & 63.0 & 78.8 & 24.7 & 48.6 & 58.7 & 75.8 \\
    \textbf{bbeh\_buggy\_tables} & 32.5 & 0.0 & 6.0 & 0.5 & 5.5 & 0.0 & 2.0 & 8.3 & 4.5 \\
    \textbf{BBH\_movie\_recommendation} & 63.2 & 42.0 & 100.0 & 28.8 & 72.4 & 10.0 & 42.0 & 36.0 & 71.5 \\
    \textbf{bbeh\_movie\_recommendation} & 33.0 & 23.5 & 83.5 & 26.5 & 84.0 & 3.5 & 16.0 & 21.6 & 61.2 \\
    \textbf{BBH\_sports\_understanding} & 74.0 & 58.0 & 100.0 & 55.6 & 99.2 & 48.4 & 95.6 & 59.0 & 98.3 \\
    \textbf{bbeh\_sportqa} & 17.0 & 35.0 & 82.0 & 19.5 & 59.5 & 4.0 & 17.5 & 18.9 & 53.0 \\
    \textbf{gsm8k} & 93.0 & 17.1 & 21.3 & 81.1 & 96.5 & 58.9 & 81.3 & 62.5 & 66.4 \\
    \textbf{MATH} & 76.7 & 42.3 & 100.0 & 40.7 & 74.7 & 39.9 & 71.1 & 49.9 & 81.9 \\
    \textbf{BBH\_multistep\_arithmetic\_two} & 94.0 & 98.8 & 100.0 & 98.4 & 99.6 & 48.4 & 82.0 & 84.9 & 93.9 \\
    \textbf{bbeh\_multistep\_arithmetic} & 6.5 & 0.0 & 0.0 & 0.5 & 7.0 & 0.0 & 3.0 & 1.8 & 3.3 \\
    \textbf{BBH\_object\_counting} & 56.0 & 82.8 & 100.0 & 86.4 & 95.6 & 48.8 & 71.2 & 68.5 & 88.9 \\
    \textbf{bbeh\_object\_counting} & 18.0 & 0.0 & 25.5 & 0.0 & 63.0 & 0.5 & 31.0 & 4.6 & 39.8 \\
    \textbf{BBH\_reasoning\_about\_colored\_objects} & 79.6 & 54.8 & 100.0 & 68.4 & 84.4 & 46.0 & 71.6 & 62.2 & 85.3 \\
    \textbf{bbeh\_object\_properties} & 21.0 & 0.5 & 25.0 & 2.0 & 37.0 & 0.0 & 22.5 & 5.9 & 28.2 \\
    \bottomrule
    \end{tabular}}
    \caption{Qwen2.5-7B Full Result.}
  \label{tab:Qwen2.5-7B}%
\end{table*}%

\begin{table*}[htbp]
  \centering
  \resizebox{!}{12cm}{%
    \begin{tabular}{crrrrrrrrr}
    \toprule
    \multirow{2}[4]{*}{\textbf{Dataset}} & \multicolumn{1}{c}{\textbf{Text}} & \multicolumn{2}{c}{\textbf{PoT}} & \multicolumn{2}{c}{\textbf{Z3}} & \multicolumn{2}{c}{\textbf{CSP}} & \multicolumn{2}{c}{\textbf{AVG}} \\
\cmidrule{2-10}          & \multicolumn{1}{c}{\textbf{ACC}} & \multicolumn{1}{c}{\textbf{ACC}} & \multicolumn{1}{c}{\textbf{Exec\_Rate}} & \multicolumn{1}{c}{\textbf{ACC}} & \multicolumn{1}{c}{\textbf{Exec\_Rate}} & \multicolumn{1}{c}{\textbf{ACC}} & \multicolumn{1}{c}{\textbf{Exec\_Rate}} & \multicolumn{1}{c}{\textbf{ACC}} & \multicolumn{1}{c}{\textbf{Exec\_Rate}} \\
    \midrule
    \textbf{Average} & 52.7 & 43.9 & 83.5 & 34.8 & 76.5 & 37.0 & 68.1 & 42.1 & 76.0 \\
    \midrule
    \textbf{FOLIO} & 83.6 & 78.4 & 99.3 & 71.6 & 100.0 & 61.2 & 97.8 & 73.7 & 99.0 \\
    \textbf{ProntoQA} & 99.0 & 84.8 & 99.4 & 54.6 & 90.2 & 81.2 & 93.8 & 79.9 & 94.5 \\
    \textbf{logicbenchBQA} & 80.1 & 70.4 & 100.0 & 72.3 & 99.1 & 64.2 & 99.6 & 71.8 & 99.6 \\
    \textbf{BoardgameQA} & 68.8 & 50.0 & 94.9 & 50.9 & 97.8 & 46.5 & 92.3 & 54.1 & 95.0 \\
    \textbf{ARLSAT} & 23.9 & 31.7 & 85.7 & 30.9 & 68.7 & 30.0 & 61.3 & 29.1 & 71.9 \\
    \textbf{BBH\_boolean\_expressions} & 96.8 & 98.8 & 100.0 & 42.0 & 74.8 & 64.4 & 99.2 & 75.5 & 91.3 \\
    \textbf{bbeh\_boolean\_expressions} & 70.0 & 0.0 & 53.0 & 3.0 & 4.0 & 22.5 & 33.0 & 23.9 & 30.0 \\
    \textbf{BBH\_formal\_fallacies} & 68.0 & 58.4 & 100.0 & 61.2 & 100.0 & 54.8 & 98.0 & 60.6 & 99.3 \\
    \textbf{bbeh\_zebra\_puzzles} & 39.0 & 12.5 & 46.5 & 13.5 & 39.5 & 8.0 & 22.0 & 18.3 & 36.0 \\
    \textbf{BBH\_logical\_deduction\_five\_objects} & 61.6 & 57.6 & 100.0 & 62.8 & 94.0 & 77.6 & 96.8 & 64.9 & 96.9 \\
    \textbf{BBH\_logical\_deduction\_seven\_objects} & 50.4 & 44.0 & 100.0 & 57.2 & 99.6 & 70.0 & 83.6 & 55.4 & 94.4 \\
    \textbf{BBH\_logical\_deduction\_three\_objects} & 84.8 & 75.6 & 100.0 & 74.8 & 91.2 & 92.0 & 94.8 & 81.8 & 95.3 \\
    \textbf{bbeh\_boardgame\_qa} & 31.5 & 0.0 & 84.5 & 12.5 & 37.0 & 18.0 & 46.0 & 15.5 & 55.8 \\
    \textbf{BBH\_tracking\_shuffled\_objects\_five\_objects} & 78.8 & 100.0 & 100.0 & 2.4 & 99.2 & 4.4 & 28.0 & 46.4 & 75.7 \\
    \textbf{BBH\_tracking\_shuffled\_objects\_seven\_objects} & 71.2 & 100.0 & 100.0 & 2.4 & 100.0 & 2.4 & 22.8 & 44.0 & 74.3 \\
    \textbf{BBH\_tracking\_shuffled\_objects\_three\_objects} & 80.8 & 99.6 & 100.0 & 8.0 & 100.0 & 5.2 & 16.8 & 48.4 & 72.3 \\
    \textbf{bbeh\_shuffled\_objects} & 42.0 & 6.0 & 19.5 & 9.5 & 23.0 & 10.5 & 23.0 & 17.0 & 21.8 \\
    \textbf{BBH\_web\_of\_lies} & 85.6 & 63.2 & 100.0 & 77.2 & 100.0 & 58.0 & 64.0 & 71.0 & 88.0 \\
    \textbf{bbeh\_web\_of\_lies} & 12.5 & 13.0 & 51.5 & 1.0 & 9.5 & 2.5 & 7.0 & 7.3 & 22.7 \\
    \textbf{bAbI15} & 99.1 & 96.1 & 97.3 & 89.6 & 99.1 & 89.2 & 98.5 & 93.5 & 98.3 \\
    \textbf{NeuLRdeductive} & 92.1 & 85.7 & 99.0 & 37.1 & 84.9 & 49.8 & 86.7 & 66.2 & 90.2 \\
    \textbf{clutrr} & 46.8 & 30.3 & 99.9 & 34.4 & 99.5 & 46.0 & 99.8 & 39.4 & 99.7 \\
    \textbf{bAbI16} & 93.0 & 96.7 & 100.0 & 78.4 & 99.6 & 83.9 & 100.0 & 88.0 & 99.9 \\
    \textbf{NeuLRinductive} & 8.2 & 8.6 & 100.0 & 18.0 & 98.6 & 14.1 & 93.8 & 12.2 & 97.5 \\
    \textbf{anli} & 84.1 & 76.7 & 100.0 & 69.7 & 100.0 & 80.3 & 99.9 & 77.7 & 100.0 \\
    \textbf{AbductionRules} & 91.0 & 85.7 & 99.8 & 47.7 & 95.0 & 48.3 & 81.1 & 68.2 & 92.0 \\
    \textbf{BBH\_causal\_judgement} & 55.1 & 61.5 & 100.0 & 55.6 & 100.0 & 49.7 & 100.0 & 55.5 & 100.0 \\
    \textbf{bbeh\_causal\_understanding} & 42.5 & 0.0 & 87.0 & 43.0 & 96.5 & 40.0 & 95.5 & 31.4 & 93.0 \\
    \textbf{NeuLRabductive} & 12.0 & 15.4 & 93.8 & 17.3 & 84.3 & 1.0 & 27.4 & 11.4 & 68.5 \\
    \textbf{logicqa} & 63.8 & 53.8 & 99.5 & 42.4 & 98.9 & 45.1 & 97.1 & 51.3 & 98.5 \\
    \textbf{BBH\_date\_understanding} & 74.0 & 54.4 & 100.0 & 40.4 & 97.2 & 55.6 & 96.8 & 56.1 & 98.0 \\
    \textbf{bbeh\_time\_arithmetic} & 32.0 & 10.5 & 59.5 & 7.5 & 43.5 & 10.5 & 45.5 & 15.1 & 49.5 \\
    \textbf{BBH\_temporal\_sequences} & 30.0 & 42.8 & 98.8 & 33.6 & 98.4 & 36.0 & 76.0 & 35.6 & 91.1 \\
    \textbf{bbeh\_temporal\_sequence} & 9.0 & 0.0 & 42.5 & 0.0 & 20.0 & 0.0 & 30.5 & 2.3 & 31.0 \\
    \textbf{BBH\_disambiguation\_qa} & 35.6 & 46.0 & 100.0 & 41.2 & 99.2 & 64.0 & 99.2 & 46.7 & 99.5 \\
    \textbf{bbeh\_disambiguation\_qa} & 42.5 & 0.0 & 43.3 & 34.2 & 90.0 & 61.7 & 100.0 & 34.6 & 77.8 \\
    \textbf{BBH\_hyperbaton} & 72.0 & 66.8 & 91.2 & 66.4 & 100.0 & 82.0 & 96.8 & 71.8 & 96.0 \\
    \textbf{bbeh\_hyperbaton} & 1.0 & 0.0 & 64.0 & 3.0 & 33.5 & 2.0 & 18.0 & 1.5 & 38.5 \\
    \textbf{BBH\_ruin\_names} & 37.2 & 32.4 & 100.0 & 28.0 & 98.8 & 39.6 & 94.4 & 34.3 & 97.7 \\
    \textbf{bbeh\_nycc} & 8.0 & 0.0 & 62.5 & 1.5 & 95.5 & 5.0 & 72.5 & 3.6 & 76.8 \\
    \textbf{BBH\_salient\_translation\_error\_detection} & 44.4 & 27.6 & 100.0 & 38.4 & 99.6 & 26.8 & 100.0 & 34.3 & 99.9 \\
    \textbf{bbeh\_linguini} & 18.0 & 0.0 & 53.5 & 2.5 & 79.0 & 10.5 & 63.0 & 7.8 & 65.2 \\
    \textbf{BBH\_snarks} & 73.6 & 50.6 & 100.0 & 51.1 & 98.9 & 41.6 & 100.0 & 54.2 & 99.6 \\
    \textbf{bbeh\_sarc\_triples} & 16.5 & 12.0 & 96.5 & 16.5 & 43.0 & 34.0 & 100.0 & 19.8 & 79.8 \\
    \textbf{BBH\_dyck\_languages} & 64.4 & 38.8 & 97.6 & 14.4 & 96.8 & 3.2 & 31.2 & 30.2 & 75.2 \\
    \textbf{bbeh\_dyck\_languages} & 2.5 & 0.0 & 24.5 & 4.5 & 47.0 & 0.5 & 18.0 & 1.9 & 29.8 \\
    \textbf{BBH\_word\_sorting} & 20.8 & 96.4 & 100.0 & 25.2 & 86.8 & 4.4 & 31.6 & 36.7 & 72.8 \\
    \textbf{bbeh\_word\_sorting} & 20.5 & 12.5 & 73.5 & 6.5 & 84.5 & 1.5 & 15.0 & 10.3 & 57.7 \\
    \textbf{BBH\_geometric\_shapes} & 53.6 & 13.6 & 100.0 & 29.2 & 99.2 & 60.8 & 99.2 & 39.3 & 99.5 \\
    \textbf{bbeh\_geometric\_shapes} & 20.0 & 1.5 & 68.5 & 17.0 & 58.5 & 42.5 & 73.0 & 20.3 & 66.7 \\
    \textbf{BBH\_navigate} & 88.8 & 75.2 & 96.0 & 65.2 & 94.0 & 55.2 & 97.2 & 71.1 & 95.7 \\
    \textbf{bbeh\_spatial\_reasoning} & 7.5 & 9.5 & 41.5 & 7.5 & 32.0 & 3.5 & 21.5 & 7.0 & 31.7 \\
    \textbf{BBH\_penguins\_in\_a\_table} & 84.3 & 98.0 & 100.0 & 82.2 & 98.6 & 50.7 & 85.6 & 78.8 & 94.8 \\
    \textbf{bbeh\_buggy\_tables} & 86.5 & 0.0 & 28.0 & 1.0 & 13.5 & 0.0 & 7.0 & 21.9 & 16.2 \\
    \textbf{BBH\_movie\_recommendation} & 75.2 & 47.2 & 99.2 & 21.2 & 90.0 & 70.8 & 100.0 & 53.6 & 96.4 \\
    \textbf{bbeh\_movie\_recommendation} & 49.0 & 0.0 & 52.5 & 11.0 & 71.0 & 9.0 & 74.5 & 17.3 & 66.0 \\
    \textbf{BBH\_sports\_understanding} & 73.6 & 56.0 & 99.2 & 62.8 & 100.0 & 53.6 & 99.2 & 61.5 & 99.5 \\
    \textbf{bbeh\_sportqa} & 11.5 & 42.5 & 87.5 & 4.0 & 12.0 & 3.0 & 11.5 & 15.3 & 37.0 \\
    \textbf{gsm8k} & 90.6 & 89.2 & 100.0 & 74.8 & 99.0 & 60.0 & 95.4 & 78.6 & 98.1 \\
    \textbf{MATH} & 71.6 & 60.8 & 92.3 & 33.1 & 68.5 & 36.4 & 85.6 & 50.5 & 82.1 \\
    \textbf{BBH\_multistep\_arithmetic\_two} & 96.4 & 98.0 & 100.0 & 95.2 & 100.0 & 85.2 & 100.0 & 93.7 & 100.0 \\
    \textbf{bbeh\_multistep\_arithmetic} & 10.5 & 0.0 & 60.5 & 0.0 & 2.0 & 0.0 & 7.0 & 2.6 & 23.2 \\
    \textbf{BBH\_object\_counting} & 41.2 & 72.8 & 100.0 & 76.0 & 91.2 & 53.6 & 88.8 & 60.9 & 93.3 \\
    \textbf{bbeh\_object\_counting} & 16.5 & 0.0 & 41.5 & 0.0 & 19.5 & 0.0 & 5.5 & 4.1 & 22.2 \\
    \textbf{BBH\_reasoning\_about\_colored\_objects} & 71.6 & 87.2 & 100.0 & 60.4 & 96.0 & 59.2 & 90.8 & 69.6 & 95.6 \\
    \textbf{bbeh\_object\_properties} & 9.5 & 1.5 & 25.0 & 1.0 & 10.0 & 0.0 & 4.0 & 3.0 & 13.0 \\
    \bottomrule
    \end{tabular}}
    \caption{Qwen2.5-7B-Base.w.Formal Full result.}
  \label{tab:Qwen2.5-7B-Base.w.Formal}%
\end{table*}%

\section{Prompts}
\begin{figure}[tp]
    \centering
    \includegraphics[width=0.5\textwidth]{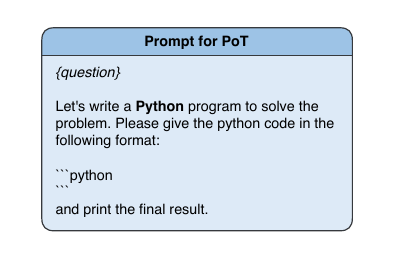}
    \caption{Prompt for PoT}
    \label{fig:Prompt for PoT}
\end{figure}

\begin{figure}[tp]
    \centering
    \includegraphics[width=0.5\textwidth]{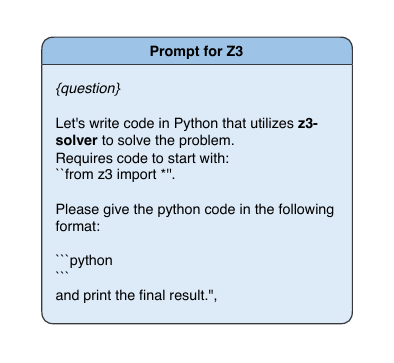}
    \caption{Prompt for Z3}
    \label{fig:Prompt for Z3}
\end{figure}

\begin{figure}[tp]
    \centering
    \includegraphics[width=0.5\textwidth]{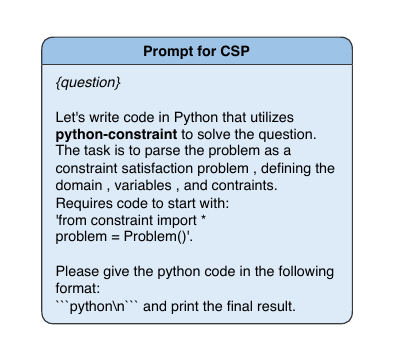}
    \caption{Prompt for CSP}
    \label{fig:Prompt for CSP}
\end{figure}

\begin{figure}[tp]
    \centering
    \includegraphics[width=0.5\textwidth]{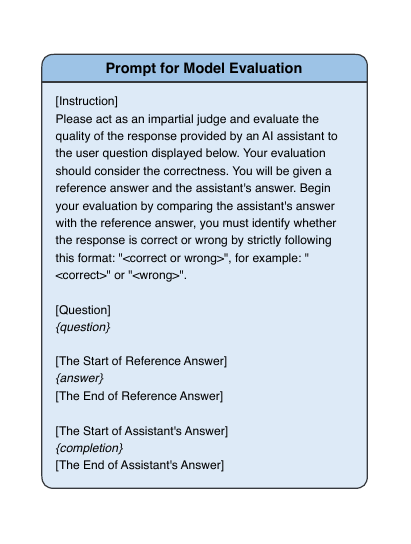}
    \caption{Prompt for Model Eval}
    \label{fig:Prompt for Model Eval}
\end{figure}

For text, we use questions directly as input to the rubric.
For formal languages, we use zero-shot reviews.
Prompts are as follows: PoT in Figure ~\ref{fig:Prompt for PoT}; Z3 in Figure ~\ref{fig:Prompt for Z3}; CSP in Figure ~\ref{fig:Prompt for CSP}.
The Prompt for evaluating models is in Figure ~\ref{fig:Prompt for Model Eval}.


\end{document}